\documentclass{article}

\usepackage[final]{neurips_2024}

\usepackage[utf8]{inputenc} %
\usepackage[T1]{fontenc}    %
\usepackage{hyperref}       %
\usepackage{url}            %
\usepackage{booktabs}       %
\usepackage{amsfonts}       %
\usepackage{nicefrac}       %
\usepackage{microtype}      %
\usepackage[dvipsnames]{xcolor}         %

\usepackage{enumitem}
\setlist[enumerate]{itemsep=0.5pt}

\hypersetup{
    colorlinks=true, %
    linkcolor=blue!60!black, %
    citecolor=blue!60!black, %
    urlcolor=blue!60!black %
}
\usepackage{bm}
\usepackage{amsmath}
\usepackage{amsfonts}
\usepackage{amssymb}
\usepackage{amsthm}
\usepackage{xspace}
\usepackage{inconsolata}
\usepackage{graphicx}
\usepackage{adjustbox}
\usepackage[normalem]{ulem}
\usepackage{float}
\usepackage{textpos}
\usepackage{listings}
\usepackage{tikz}
\usetikzlibrary{shapes, arrows.meta, positioning}
\usepackage[capitalize]{cleveref}
\crefname{lstfloat}{listing}{listings}
\Crefname{lstfloat}{Lst.}{Lsts.}
\Crefname{equation}{Eq.}{Eqs.}
\Crefname{figure}{Fig.}{Figs.}
\Crefname{tabular}{Tab.}{Tabs.}
\Crefname{appendix}{App.}{Apps.}
\Crefname{section}{Sec.}{Secs.}

\theoremstyle{plain}

\theoremstyle{definition}

\usepackage{tagging}

\title{Universal In-Context Approximation\\By Prompting Fully Recurrent Models}

\author{%
  Aleksandar Petrov, ~Tom A. Lamb, ~Alasdair Paren, ~Philip H.S. Torr, ~Adel Bibi\\
  Department of Engineering Science\\
  University of Oxford \\
  \texttt{aleks@robots.ox.ac.uk}
}

\newcommand*{\smalldots}{.\kern-0.02em.\kern-0.02em.}

\newcommand{\RR}{\mathbb{R}}

\newcommand{\XX}{\mathcal{X}}
\newcommand{\YY}{\mathcal{Y}}

\newcommand{\CC}{\mathcal{C}}
\newcommand{\CCtok}{\mathcal{C}^\text{tok}}
\newcommand{\CCvec}{\mathcal{C}^\text{vec}}

\newcommand{\HH}{\mathcal{H}}
\newcommand{\HHtok}{\mathcal{H}^\text{tok}}

\newcommand{\HHvec}{\mathcal{H}^\text{vec}}

\newcommand{\Input}{\texttt{Input}\xspace}
\newcommand{\ReLU}{\texttt{ReLU}\xspace}
\newcommand{\Lin}{\texttt{Lin}\xspace}
\newcommand{\LinState}{\texttt{LinState}\xspace}
\newcommand{\Multi}{\texttt{Multi}\xspace}
\newcommand{\ForEach}{\texttt{ForEach}\xspace}
\newcommand{\Slice}{\texttt{Slice}\xspace}
\newcommand{\Concat}{\texttt{Concat}\xspace}

\DeclareMathOperator{\Mod}{mod}

\newcommand{\const}[1]{#1}

\begin{document}

\newfloat{lstfloat}{htbp}{lop}
\floatname{lstfloat}{Listing}
\def\lstfloatautorefname{Listing} %

\maketitle

\makeatletter
\lst@AddToHook{OnEmptyLine}{\vspace{-0.6\baselineskip}\addtocounter{lstnumber}{-1}}
\makeatother

\begin{abstract}
    Zero-shot and in-context learning enable solving tasks without model fine-tuning, making them essential for developing generative model solutions.
    Therefore, it is crucial to understand whether a pretrained model can be prompted to approximate any function, i.e., whether it is a universal in-context approximator.
    While it was recently shown that transformer models do possess this property, these results rely on their attention mechanism.
    Hence, these findings do not apply to fully recurrent architectures like RNNs, LSTMs, and the increasingly popular SSMs.
    We demonstrate that RNNs, LSTMs, GRUs, Linear RNNs, and linear gated architectures such as Mamba and Hawk/Griffin can also serve as universal in-context approximators.
    To streamline our argument, we introduce a programming language called LSRL that compiles to these fully recurrent architectures.
    LSRL may be of independent interest for further studies of fully recurrent models, such as constructing interpretability benchmarks.
    We also study the role of multiplicative gating and observe that architectures incorporating such gating (e.g., LSTMs, GRUs, Hawk/Griffin) can implement certain operations more stably, making them more viable candidates for practical in-context universal approximation.
\end{abstract}

\section{Introduction}
Until recently, solving a task with machine learning required training or fine-tuning a model on a dataset matching the task at hand.
However, large foundation models exhibit the ability to solve new tasks without being specifically fine-tuned or trained for them: often it is sufficient to simply prompt them in the right way.
\untagged{neurips}{This has made prompting a key method for steering a model towards a specific behaviour or task \citep{liu2023pretrain}.}
Prompting has been especially successful because of \emph{in-context learning}: the ability to modify the model's behavior with information provided within the input sequence, without changing the underlying model parameters \citep{brown2020language}.
\untagged{neurips}{As a result, the art and skill of constructing a successful prompt (\emph{prompt engineering}) has become extremely important \citep{liu2022design,sahoo2024systematic}.}
Yet, we know little about the theoretical properties of prompting.
It is not even clear if there are limits to what can be achieved with prompting or, conversely, whether it is possible to prompt your way into any behaviour or task.

This can be framed as a universal approximation question.
Classically, universal approximation results show how a class of tractable functions, such as neural networks, approximates another class of concept functions, e.g., all continuous functions on a bounded domain, with arbitrary accuracy.
This is often done by showing that one can choose \emph{model parameters} that approximate the target function.
However, in-context learning poses a different challenge as the model parameters are \emph{fixed}.
Instead, a part of the input (the prompt) is modified to cause the model to approximate the target function.
Hence, we define universal \emph{in-context} approximation to be the property that there exist fixed weights such that the resulting model can be prompted to approximate any function from a concept class.
Understanding whether a model can be a universal \emph{in-context} approximator is especially important as most commercial models are accessible exclusively via a prompting interface \citep{lamalfa2023language}.

In-context learning has been almost exclusively studied in conjunction with the transformer architecture \citep{vaswani2017attention}.
This is likely because in-context abilities appear once the models are large enough \citep{wei2021finetuned} and most large models have been transformer-based.
On the subject of universal in-context approximation, \citet{wang2024universality} were first to show that a transformer possesses this property by discretising and memorising all possible functions in the model weights.
Memorisation is not needed, though, and even small transformers can be universal approximators when prompted \citet{petrov2024universal}.
Both results, however, critically depend on the attention mechanism of the transformer architecture \citep{bahdanau2014neural}.

Still, generative models are not restricted to attention-based architectures: there are the ``classic'' recurrent neural networks (RNNs, \citealp{amari1992rnns}), long short-term memory models (LSTMs, \citealp{hochreiter1997long}) and gated recurrent units (GRUs, \citealp{cho2014learning}). 
Recently, Linear RNN models (also known as state-space models or SSMs) were proposed as a scalable alternative to the transformer architecture \citep{orvieto2023resurrecting,fu2023hungry} and have started to outperform
similarly-sized transformers when multiplicative gating is added \citep{gu2023mamba,de2024griffin,botev2024recurrentgemma}. %
Furthermore, despite in-context learning being associated with the transformer, recent empirical results show in-context learning in SSMs, RNNs, LSTMs and even convolutional models \citep{xie2021explanation,akyurek2024context,lee2023exploring}. 

Yet, despite their ability to be in-context learners, there is little known about the theoretical properties of these fully recurrent architectures.
As these architectures become more and more widely used, understanding their in-context approximation abilities is increasingly more important for their safety, security and alignment. 
We show that, in fact, many of these architectures, similarly to transformers, can be universal in-context approximators.
Concretely, our contributions are as follows:

\begin{enumerate}
    \item We develop \emph{Linear State Recurrent Language} (LSRL): a programming language that compiles to different fully recurrent models.
    Programming in LSRL is akin to ``thinking like a recurrent model''. LSRL programs can then be implemented exactly as model weights.
    \item Using LSRL, we construct Linear RNN models that can be prompted to act as any token-to-token function over finite token sequences, or to approximate any continuous function. These results also hold for RNNs, LSTMs, GRUs and Hawk/Griffin models \citep{de2024griffin}.
    \item We present constructions with and without multiplicative gating.
    However, we observe that the constructions without these gates depend on numerically unstable conditional logic.
    \item Nevertheless, we show that multiplicative gates lead to more compact and numerically stable models, making it more likely that universal in-context approximation properties arise in models utilising them, such as LSTMs, GRUs and the latest generation of Linear RNNs.
\end{enumerate}

\section{Preliminaries}
\label{sec:preliminaries}

\vspace{-0.7em}
\paragraph{Fully recurrent architectures.}
In this work, we focus exclusively on fully recurrent neural network architectures.
Recurrent models operate over sequences.
Concretely, consider an input sequence $(\bm x_1, \ldots, \bm x_N)$ with $\bm x_t\in\XX$, $\XX$ being some input space.
We will refer to the elements of the input sequence as \emph{tokens} even if they are real-valued vectors.
A recurrent model $g: \XX^\star \to \YY$ maps a sequence of inputs to an output in some output space $\YY$.
These models are always causal, namely:
\begin{equation}
    \bm y_t = g(\bm x_1, \ldots, \bm x_t). \label{eq:recurrent_models}
\end{equation}
We will abuse the notation and refer to $(\bm y_1, \smalldots, \bm y_t) {=} (g(\bm x_1), \smalldots, g(\bm x_1, \smalldots, \bm x_t))$ as simply $g(\bm x_1, \smalldots, \bm x_t)$.
We will also separate the input sequence into a query $(\bm q_1, \smalldots, \bm q_n)$ and a prompt $(\bm p_1, \smalldots, \bm p_N)$.
The prompt specifies the target function $f$ that we approximate while the query designates the input at which we evaluate it.
Contrary to the typical setting, we will place the query before the prompt.%
\footnote{That is necessitated by the limited capacity of the state variables.
As the model is fixed, in order to increase the precision of the approximation, we can only increase the prompt length.
If the prompt is before the query, it would have to be compressed into a fixed-size state, limiting the approximation precision even with increased prompt lengths.
But if the query has a fixed size, it can be stored in a fixed-size state variable exactly.}

There are various neural network architectures that fall under the general framework of \Cref{eq:recurrent_models}.
The quintessential one is the RNN.
It processes inputs one by one with only a non-linear state being passed from one time step to the other.
A model $g$ can thus be stacked RNN layers, each one being:
\begin{align}
    \begin{aligned}
        \bm s_t &= \sigma(\bm A \bm s_{t-1} + \bm B \bm x_t + \bm b), \\
        \bm y_t   &= \phi(\bm s_t),
    \end{aligned}
    &&\text{(Classic RNN)}
    \label{eq:rnn}
\end{align}
with $\bm A, \bm B, \bm b$ and the initial state value $\bm s_0$ being model parameters, $\sigma$ a non-linear activation function and $\phi$ a multi-layer perceptron (MLP) with \ReLU activations.
We assume that $\sigma$ is always a \ReLU to keep the analysis simpler.
The non-linearity in the state update can make the model difficult to train (vanishing and exploding gradients, \citealp{bengio1994learning}).
Therefore, Linear RNNs have been proposed as regularizing the eigenvalues of $\bm A$ can stabilise the training dynamics \citep{orvieto2023resurrecting}. 
Linear RNNs also admit a convolutional representation, making them trainable in parallel \citep{gu2021efficiently,fu2023hungry}.
Linear RNNs drop the non-linearity from the state update in \Cref{eq:rnn}:
\begin{align}
    \begin{aligned}
        \bm s_t &= \bm A \bm s_{t-1} + \bm B \bm x_t + \bm b, \\
        \bm y_t   &= \phi(\bm s_t).
    \end{aligned}
    &&\text{(Linear RNN)}
    \label{eq:linear_rnn}
\end{align}
The fully linear state updates do not affect the expressivity of the models, as non-linear activations are nevertheless present in the MLP layers $\phi$ between the linear state update layers \citep{wang2024state,boyd1985fading}.
The state-of-the-art Linear RNN models also utilise some form of multiplicative gating \citep{gu2023mamba,de2024griffin,botev2024recurrentgemma}.
While specific implementations can differ, we can abstract it as the following Gated Linear RNN architecture:
\begin{align}
    \begin{aligned}
        \bm s_t &= \bm A \bm s_{t-1} + \bm B \bm x_t + \bm b, \\
        \bm y_t   &= \gamma(\bm x_t) \odot \phi(\bm s_t),
    \end{aligned}
    &&\text{(Gated Linear RNN)}
    \label{eq:gated_linear_rnn}
\end{align}
with $\gamma$ being another MLP and $\odot$ being the element-wise multiplication operation (Hadamard product).
\Cref{eq:gated_linear_rnn} encompasses a range of recently proposed models.
For example, one can show that any model consisting of $L$ stacked Gated Linear RNN layers, with $\gamma$ and $\phi$ with $k$ layers, can be represented as a $L(k{+}2)$-layer Hawk or Griffin model \citep{de2024griffin}.
The conversions are described in detail in \Cref{sec:hawk_griffin}.
We can similarly add multiplicative gating to the classic RNN architecture:
\begin{align}
    \begin{aligned}
        \bm s_t &= \sigma(\bm A \bm s_{t-1} + \bm B \bm x_t + \bm b), \\
        \bm y_t &= \gamma(\bm x_t) \odot \phi(\bm s_t),
    \end{aligned}
    &&\text{(Gated RNN)}
    \label{eq:gated_rnn}
\end{align}
\Cref{eq:gated_rnn} may appear unusual but it is related to the well-known GRU \citep{cho2014learning} and LSTM \citep{hochreiter1997long} architectures.
Same as the case with Griffin/Hawk, any Gated RNN can be represented as a $L(k{+}2)$-layer GRU or LSTM model (details in \Cref{sec:GRU,sec:lstms}).
As a result, if there exists a Gated RNN model that is a universal in-context approximator (which we later show to be the case), then there also exist GRU and LSTM models with the same property.

\untagged{neurips}{%
All the models above can be boiled down to compositions of a few building blocks.
Namely, linear layers, \ReLU activations, (non-)linear state updates and multiplicative operations (in the case of gated models).
These four building blocks will be the primitives of LSRL, the programming language we introduce in \Cref{sec:lsrl} as a tool to write programs that directly compile to these architectures.
In practice, a number of additional elements might be present such as residual connections \citep{he2016deep}, positional embeddings \citep{su2024roformer} and normalisation layers \citep{ba2016layer,zhang2019root}. 
However, as these are not necessary for showing the in-context universal approximation abilities of the four architectures above, we will not consider them in this work.}

\vspace{-0.7em}
\paragraph{Theoretical understanding of in-context learning.}
Beyond the question of universal in-context approximation, there have been attempts to theoretically understand in-context learning from various perspectives. 
The ability to learn linear functions and perform optimization in-context has been extensively explored in the context of linear regression \citep{garg2022can,akyurek2022learning,oswald23transformers,fu2023transformers,zhang2023trained,ahn2023transformers}, kernel regression \citep{han2023context} and dynamical systems \citep{li2023transformers}.
Furthermore, studies have explored how in-context learning identifies and applies the appropriate pretraining skill \citep{xie2021explanation,coda2023meta,bai2024transformers}.
It has also been shown that transformers can construct internal learning objectives and optimize them during the forward pass \citep{von2023uncovering,dai2023gpt}. 
However, these studies almost exclusively focus on the transformer architecture, and the applicability of their findings to fully recurrent models remains unclear.

\vspace{-0.7em}
\paragraph{Approximation theory.}
Let $\XX$ and $\YY$ be normed vector spaces.
Take a set of functions $\CC\subseteq\YY^\XX$ from $\XX$ to $\YY$ called a \emph{concept space}.
Take also a set of nicely behaved functions $\HH\subset\YY^\XX$, called \emph{hypothesis space}. 
$\HH$ could be any set that we have tools to construct and analyse, e.g., all polynomials or all neural networks of a particular architectural type.
Approximation theory is concerned with how well functions in $\HH$ approximate functions in $\CC$.
We say that $\HH$ \emph{universally approximates} $\CC$ over a compact domain $\mathcal D$ (or that \emph{$\HH$ is dense in $\CC$}) if for every $f{\in}\CC$ and $\epsilon{>}0$ there exist a $h{\in}\HH$ such that $\sup_{\bm x\in\mathcal  D} |f(\bm x)\texttt{-}h(\bm x)| {\leq}\epsilon$.
There is a long history of studying the concept class of continuous functions and hypothesis classes of single hidden layer neural networks \citep{cybenko1989approximation,barron1993universal} or deeper models \citep{hornik1989multilayer,telgarsky2015representation}.
The concept class of sequence-to-sequence functions has been shown to be universally approximated with the hypothesis classes of transformers \citep{yun2019transformers}, RNNs \citep{schafer2006recurrent} and Linear RNNs \citep{wang2024state}.

The hypothesis spaces in this work are different.
The model is fixed and only the prompt part of the input is changed, i.e., all learnable parameters are in the prompt.
Take a recurrent model $g$ as in \Cref{eq:recurrent_models} with \emph{fixed} model parameters and a query length $n$. 
The  hypothesis class is all functions that result by calling $g$ on the user query followed by the prompt and taking the last $n'$ outputs:
\begin{equation}
    \HH_g^{\mathcal{D}^n} = \{ (\bm q_1, \ldots, \bm q_n) \mapsto g(\bm q_1, \ldots, \bm q_n, \bm p_1, \ldots, \bm p_N)[\texttt{-}n'\texttt{:}] \mid \forall \bm p_i\in\mathcal{D}, N>0  \}.
    \label{eq:hypothesis class}
\end{equation}
The domain $\mathcal{D}$ of $\bm p_i$ and $\bm q_i$ can be continuous embeddings in $\RR^d$ or discrete tokens $\mathcal V = \{1,\smalldots,V\}$.

Note that each $h{\in}\HH_g$ is identified by a prompt $(\bm p_1, \smalldots, \bm p_N)$ but is a function with domain all possible queries $(\bm q_1, \smalldots, \bm q_n)$.
Therefore, finding a hypothesis $h{\in}\HH_g$ that approximates a target function $f$ is equivalent to finding the prompt of that hypothesis.
The approximation properties of $\HH_g$ in \Cref{eq:hypothesis class} depend on the architecture of $g$, as well as its specific parameters.
\untagged{neurips}{%
This makes it challenging to do approximation in the context window.
The possibilities for interaction between the inputs are limited and the effects of the fixed model weights can be difficult to study \citep{petrov2023prompting}.
To the best of our knowledge, this has only been studied in the case where $g$ is a transformer model.
\citet{wang2024universality} showed that in-context universal approximation is possible with a transformer by discretizing and memorising all possible functions in the model weights, while, \citep{petrov2024universal} argues that no memorisation is needed and that a transformer with $n{+}2$ layers can be a universal approximator for sequence-to-sequence functions with input length $n$ with a prompt of length $\mathcal O(\epsilon^{-10-14d-4d^2})$.}

We study the recurrent architectures in \Cref{eq:rnn,eq:linear_rnn,eq:gated_linear_rnn,eq:gated_rnn} and their ability to approximate continuous functions over real-valued vectors and to represent discrete maps over tokens (which corresponds to how language models are used in practice).
We consider the following classes of functions.
$\CCvec{=}(\RR^{d_\text{out}}) ^ {[0,1]^{d_\text{in}}}$ contains all continuous functions from the unit hypercube to $\RR^{d_\text{out}}$, while $\CCtok{=}\{h {\in} (\mathcal V^l)^{\mathcal  V^l} \mid h \text{ causal}\}$ all causal functions from $l$ tokens to $l$ tokens.
The hypothesis classes are $\HHvec(g)$ corresponding to \Cref{eq:hypothesis class} with $D{=}[0,1]^{d_\text{in}}, n{=}n'{=}1$ and $g$  some \emph{fixed} model of one of the four architectures in \Cref{eq:rnn,eq:linear_rnn,eq:gated_linear_rnn,eq:gated_rnn}, and $\HHtok(g)$ with $D{=}\mathcal V$ and $n{=}n'{=}l$.

\section{Linear State Recurrent Language (LSRL)}
\label{sec:lsrl}

\begin{figure}
    \centering
    \includegraphics[width=\linewidth]{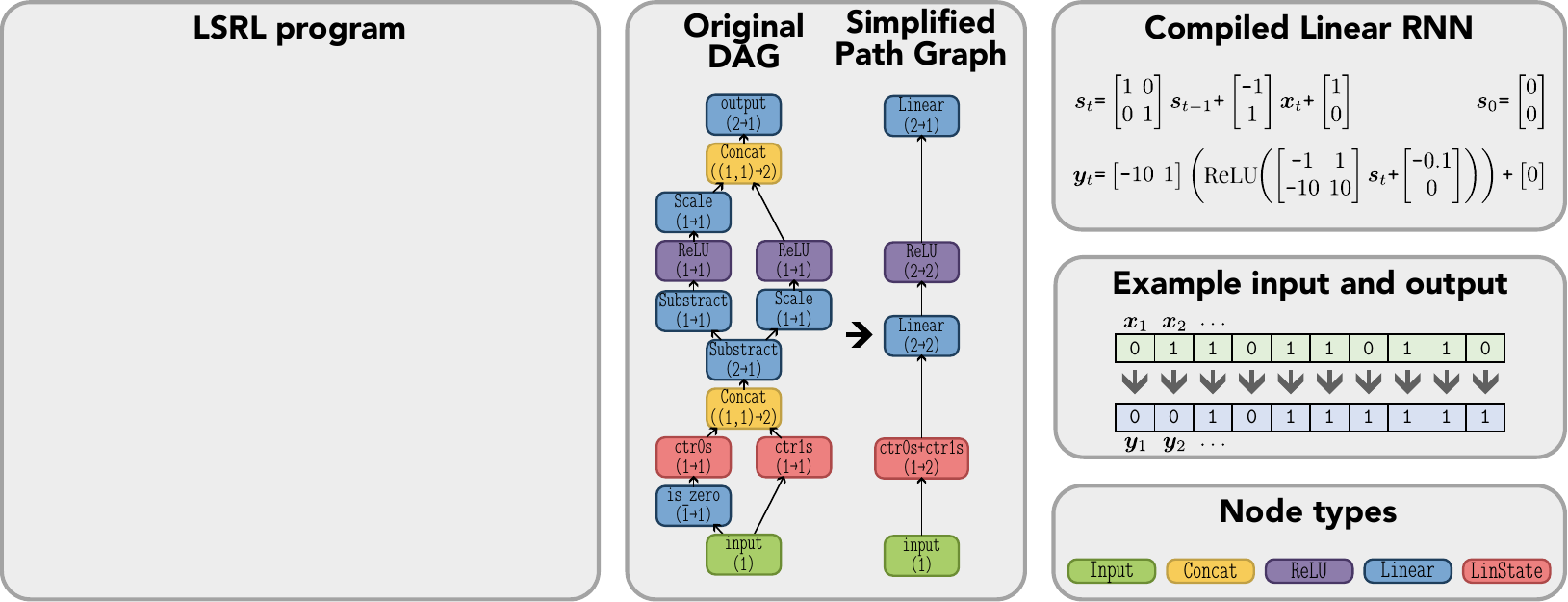}
    \caption{
    \textbf{Compilation of an LSRL program to a Linear RNN.}
    An example of a simple LSRL program that takes a sequence of 0s and 1s as an input and outputs 1 if there have been more 1s than 0s and 0 otherwise.
    The LSRL compiler follows the rules in \Cref{sec:debranching_rules} to simplify the computation DAG into a path graph.
    The resulting path graph can be represented as a Linear RNN with one layer.}
    \label{fig:basic_example}
\begin{textblock*}{7cm}(0.4cm, -6.8cm) %
        \begin{minipage}{\textwidth}
            \begin{adjustbox}{max width=\textwidth, max height=2.3cm} %
                \begin{lstlisting}[basicstyle=\ttfamily\footnotesize,frame=none]
ForEach:
input = Input(dim=1)

ctr1s=LinState(input, 
  A=[[1]], B=[[1]],
  init_state=[[0]])

# equivalent to 1-input
is_zero = f_not(input) 
ctr0s = LinState( is_zero, 
  A=[[1]], B=[[1]],
  init_state=[[0]])

# equiv. to f_step(ctr1s-ctr0s)
output=f_larger(
  ctr1s,ctr0s,mu=10)
return output
                \end{lstlisting}
            \end{adjustbox}
        \end{minipage}
\end{textblock*}
\end{figure}

We can construct the weights for universal in-context models with the architectures in \Cref{eq:rnn,eq:linear_rnn,eq:gated_linear_rnn,eq:gated_rnn} by hand but this is labour-intensive, error-prone, difficult to interpret, and the specific weights would be architecture-dependent.
Working at such a low level of abstraction can also obfuscate common mechanisms and design patterns, making it more difficult to appreciate both the capabilities and the constraints of fully recurrent architectures.
Instead, 
we propose a new programming language: \emph{Linear State Recurrent Language} (LSRL).\footnote{Our implementation of LSRL is available  at \url{https://github.com/AleksandarPetrov/LSRL}}
LSRL programs compile to the four architectures in \Cref{eq:rnn,eq:linear_rnn,eq:gated_linear_rnn,eq:gated_rnn}.
Conversely, any Linear RNN can be represented as an LSRL program, making LSRL a versatile tool for studying the capabilities of recurrent models.
Later, in \Cref{sec:ua_linear_RNNs,sec:ua_gated_linear_RNNs,sec:ua_nonlinear_RNNs} we make use of LSRL to develop programs that are universal approximators for $\CCvec$ and $\CCtok$, 
thus showing that all four architectures can be universal in-context approximators.

\vspace{-0.7em}
\paragraph{LSRL syntax.}
An LSRL program specifies how a single element is processed and how the recurrent states are updated for the next element.
LSRL programs always start with an $\Input(\bm x)=\bm x$ with an $\bm x$ of a fixed dimension.
Only one \Input can be declared in a program.
Linear layers and \ReLU{}s are also supported: $\Lin[\const{\bm A},\const{\bm b}](\bm x) := \const{\bm A} \bm x + \const{\bm b}$, $\ReLU(\bm x):=\max(\bm 0, \bm x)$.
The unique component of LSRL, however, is its $\LinState$ operation implementing the linear state update in Linear RNNs (\Cref{eq:linear_rnn}): $\LinState[\bm A, \bm B, \bm b, \bm s_0](\bm x_t) := \bm A \bm s_{t-1} + \bm B \bm x_t + \bm b $, 
where the state $\bm s_{t-1}$ is the output of the call this node at step $t-1$.
\LinState is the only way information can be passed from previous tokens to the current one.
We also provide a \Concat operation that combines variables: $\Concat(\bm x, \bm y) := (\bm x_1, \smalldots, \bm x_{|\bm x|}, \bm y_1, \smalldots, \bm y_{|\bm y|})$. 
Finally, to support gating architectures we also implement a rudimentary \Multi operation that splits its input into two sub-arrays and returns their element-wise multiplication: $\Multi(\bm x) := \bm x\texttt{[}: \nicefrac{| \bm x |}{2} \texttt{]} \odot \bm x\texttt{[} \nicefrac{| \bm x |}{2} :\texttt{]}$.
Naturally, \Multi requires that $\bm x$ has even length.
These six operations can be composed into a direct acyclic graph (DAG) with a single source node (the \Input variable) and a single sink node (marked with a \texttt{return} statement).

Such a program operates over a single token $\bm x_t$ passed to \Input, while a recurrent model needs to operate over sequences.
Thus, we wrap the program into a \ForEach loop that passes each element individually for the DAG to output a variable denoted by a \texttt{return} clause.
Each element is processed by the exact same program, with the only difference being that the state of the \LinState variables is changing between iterations.
You can see an example of a small LSRL program in \Cref{fig:basic_example}.

\vspace{-0.7em}
\paragraph{Expressiveness limitations.}
\ForEach does not behave like the typical \texttt{for} loop: only the states are accessible between iterations, i.e., you cannot use the output of a linear layer at step $t$ in any computation at step $t+1$.
Furthermore, as the program is a DAG and only states of \LinState nodes are passed between iterations, variables computed in latter operations of a previous time step are not accessible as inputs in earlier layers (with respect to the topological sorting of the computation graph).
This leads to a key programming paradigm in LSRL: a \LinState update cannot depend non-linearly on its own state.
That includes it depending on a variable that depends on the \LinState itself and conditional updates to the state.
Such a dependency would break the DAG property of the program.\footnote{For example, we cannot implement an operation that adds one to the state and squares it at each time step: $s_{t+1}=(s_t+1)^2$ or an operation that performs conditional assignment $s_{t+1} = 0 \texttt{ if } (s_t > 5) \texttt{ else } s_t$.} 
This poses serious limitations on what algorithms can be expressed in a Linear RNN and makes programming them challenging.
Still, in \Cref{sec:ua_linear_RNNs} we show how carefully constructing state updates and auxiliary variables can nevertheless allow to program some limited conditional behaviours.

\vspace{-0.7em}
\paragraph{Compilation.}
Any LSRL program without \Multi nodes can be compiled to a Linear RNN (\Cref{eq:linear_rnn}) or to a Gated Linear RNN (\Cref{eq:gated_linear_rnn}).
If the program has \Multi nodes, then it cannot be compiled to a Linear RNN as the multiplicative gating cannot be implemented exactly.
However, it can be compiled to a Gated Linear RNN.
To compile an LSRL program to a Linear (Gated) RNN, we first parse the program to build a computation graph.
This is a DAG with a single source (the \Input node) and a single sink (the \texttt{return} statement of the \ForEach loop).
At the same time, a Linear (Gated) RNN can be represented as a path graph (no branching) with the six basic operations as nodes. 
Therefore, the compilation step needs to transform this DAG into a path graph.
We achieve that by iterativly collapsing the first branching point into a single node.
The exact rules that achieve that are described in \Cref{sec:debranching_rules}. 
Later, in \Cref{sec:ua_nonlinear_RNNs}, we will show how any Linear (Gated) RNN can be converted into a \emph{non-linear} (Gated) RNN, hence, how we can compile LSRL programs to these architectures as well.

\begin{lstfloat}[t]
\begin{lstlisting}
ForEach:
input = Input(dim=1+d_in+d_out)

# counter needed to know whether we are looking at the query or the prompt
const_1 = f_constant(input, 1)
counter_vector = LinState(input=const_1, A=ones(d_in,d_in), B=ones(d_in,1), init_state=zeros(d_in,1))

# copy the query in a state (only when the counter is 1)
q_update = f_ifelse(cond=f_smaller(counter_vector, 1.5), t=input[: d_in], f=input[: d_in]*0)
q = LinState(input=q_update, A=eye(d_in), B=eye(d_in), init_state=zeros(d_in,1))

# the following operations will only change the output when counter > 1
# the step size is the first element of every prompt element
step_size = Linear(input=input[0], A=ones(d_in,1), b=zeros(d_in,1))

# using it we can compute the upper bounds of the current prompt cell
lb = input[1 : 1 + d_in]
ub = lb + step_size

# now check if q is in this cell (the bump should be 1 on all dimensions)
q_in_bump_componentwise = f_bump(q, lb, ub)
bump_sum = Linear(input=q_in_bump_componentwise, A=ones(1,d_in), b=zeros(1,1))
in_cell = f_larger(bump_sum, d_in - 0.5)
in_and_processing = f_and(in_cell, f_larger(counter, 0.5))

# if counter>1 and this cell contains q, add the value to the output state
update = f_ifelse(cond=f_larger(in_and_processing,0.5), t=input[-d_out:], f=input[-d_out:]*0)
y = LinState(input=update, A=eye(d_out), B=eye(d_out), init_state=zeros(d_out,1))
return y
\end{lstlisting}
\caption{\textbf{LSRL program for universal approximation in-context for continuous functions.} The inputs are $\bm q = [\bm q'^\top, \bm 0_{d_\text{out}+1}^\top]^\top$ with $\bm q' \in [0,1]^{d_\text{in}}$ being the query value at which we want to evaluate the function, then followed by prompts describing the target function as in \Cref{eq:continous_prompt_def}.}
\label{lst:continous_approximation}
\end{lstfloat}

\vspace{-0.7em}
\paragraph{Syntactic sugar.}
To make programming easier, we define several convenience functions.
For instance, we can \Slice variables $\bm x[\const{l}{:}\const{u}]$ via sparse \Lin layers. 
We can also sum variables and element-wise multiplication with scalars (implemented as \Lin layers).
For logical operations we also need step functions which can be approximated with \ReLU{}s: $ \texttt{f\_step}[\const{\mu}](\bm x) :=  \ReLU(\const{\mu} \bm x) - \const{\mu} \ReLU(\bm x - \const{\nicefrac{1}{\mu}})$, where $\const{\mu}$ is a positive constant controlling the quality of the approximation.
We can also approximate bump functions (1 between $l$ and $u$ and 0 otherwise): $ \texttt{f\_bump}[\bm l, \bm u, \const{\mu}](\bm x) := \texttt{f\_step}[\const{\mu}](\bm x - \bm l) - \texttt{f\_step}[\const{\mu}](\bm x - \bm u)$.
Similarly, we can approximate conjunction (\texttt{f\_and}), disjunction (\texttt{f\_or}), negation (\texttt{f\_not}), and comparison operators (\texttt{f\_larger} and \texttt{f\_smaller}).
See \Cref{sec:sugar_in_lsrl} for the definitions.

Critically, we need also a conditional operator that assigns a value $\texttt{t}(\bm x)$ if a certain condition is met and another value $\texttt{f}(\bm x)$ otherwise.
One way to implement this is:
\begin{equation}
    \begin{aligned}
    \texttt{f\_ifelse}[\texttt{cond}, \texttt{t}, \texttt{f}, \const{\lambda}](\bm x)
    &:=
    \ReLU(\texttt{-}\const{\lambda}\, \texttt{cond}(\bm x) \texttt{+} \texttt{f}(\bm x))
    + \ReLU(\texttt{-}\const{\lambda}\, \texttt{f\_not}(\texttt{cond}(\bm x)) \texttt{+} \texttt{t}(\bm x)) \\
    &\hspace{0.35em}- \ReLU(\texttt{-}\const{\lambda}\, \texttt{cond}(\bm x) \texttt{-} \texttt{f}(\bm x))
    - \ReLU(\texttt{-}\const{\lambda}\, \texttt{f\_not}(\texttt{cond}(\bm x))  \texttt{-} \texttt{t}(\bm x)),
    \end{aligned}
    \label{eq:ifelse_direct}
\end{equation}
where $\const{\lambda}$ is a constant that is larger than any absolute value that $\texttt{t}(\bm x)$ and $\texttt{f}(\bm x)$ can attain.
This construction, however, is not numerically stable \untagged{neurips}{(consider if $\texttt{cond}(\bm x)$ is not exactly 0 but a small positive number)} and we will study alternatives in \Cref{sec:ua_gated_linear_RNNs}.
We provide both numerical (\texttt{SciPy.sparse}, \citealt{SciPy}) and symbolic (\texttt{SymPy}, \citealt{SymPy}) backends with the second being crucial for programs that are not numerically stable.

\untagged{neurips}{%
\vspace{-0.7em}
\paragraph{Constant and dynamic variables.}
It is important also to distinguish between variables which can be dynamically assigned and such that must by ``baked in'' the model weights and be constant.
Some operations can only be performed when one of the operands is a constant.
For example, with a Linear RNN we cannot exactly compute the product of two variables ---such as $\Lin[\const{\bm A_1},\const{\bm b_1}](\bm x) \odot \Lin[\const{\bm A_2},\const{\bm b_2}](\bm x)$--- but we can compute a product with a fixed vector $\bm v \odot \Lin[\const{\bm A_2},\const{\bm b_2}](\bm x) $.
This is also why $\lambda$ in \Cref{eq:ifelse_direct} cannot be dynamically computed depending on the input $\bm x$.
This is not the case for the gated architectures, where variable product is possible, something we will leverage to construct more numerically stable conditional operators in \Cref{sec:ua_gated_linear_RNNs}.}

\vspace{-0.7em}
\paragraph{Prior work on encoding algorithms in model weights.}
A similar approach to developing a programming language that compiles to model weights was already done for the transformer architecture with the RASP language \citep{weiss2021thinking} and the Tracr compiler \citep{lindner2024tracr}.
They were predominantly created as a tool for interpretability research. 
In a sense, RASP is to a transformer as LSRL is to a (Linear) (Gated) RNN.
Hence, can be used to develop benchmarks for interpretability methods for fully-recurrent architectures.
However, while RASP can only express a subset of transformer models, LSRL is isomorphic to the set of all (Gated) Linear RNNs (though not to the non-linear ones).
That means that any (Gated) Linear RNN can be represented and analysed as an LSRL program and vice versa. 
Hence, the limitations of what you can express in LSRL are also limitations of what a Linear (Gated) RNN can do.
Namely: (\emph{i}) we cannot have exact multiplicative interactions between inputs without multiplicative gates, and (\emph{ii}) we cannot have state variable updates depending non-linearly on their previous iterations or in any way on a variable that depends on them.

\section{Universal In-Context Approximation with Linear RNNs}
\label{sec:ua_linear_RNNs}

\begin{figure}
    \centering
    \includegraphics[width=0.9\textwidth]{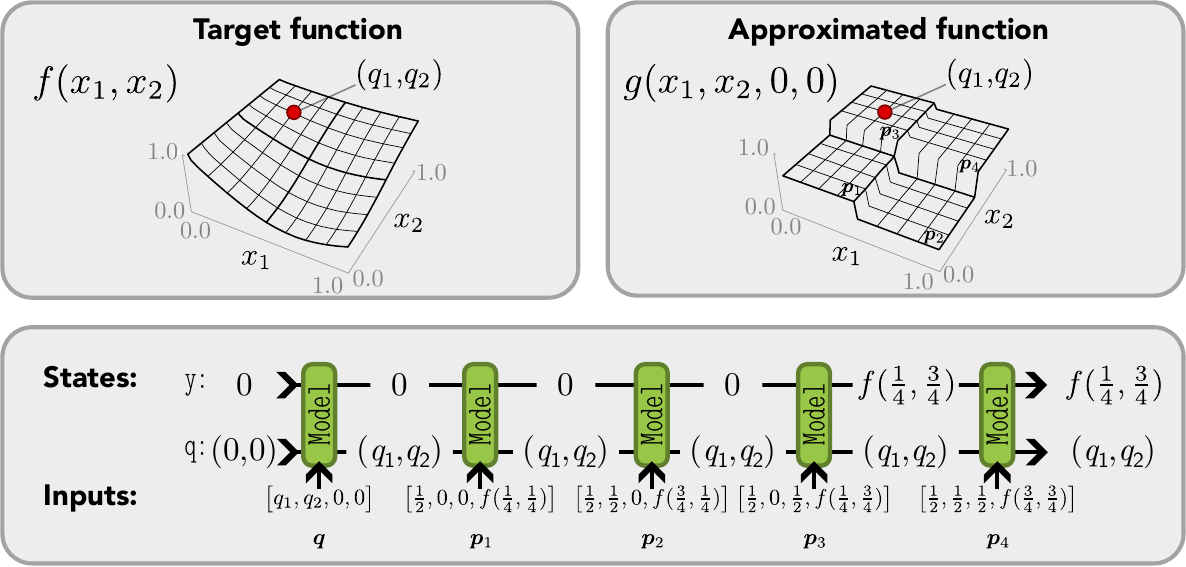}
    \caption{\textbf{Intuition behind the LSRL program for universal in-context approximation for continuous functions in \Cref{lst:continous_approximation}.} Our target function $f$ has input dimension $d_\text{in}=2$ and output dimension $d_\text{out}=1$. Each input dimension is split into two parts, hence $\delta=\nicefrac{1}{2}$. We illustrated an example input sequence of length 5: one for the query and four for the prompt tokens corresponding to each of the discretisation cells. The query $(q_1, q_2)$ falls in the cell corresponding to the third prompt token. We show how the two \LinState variables in the program are updated after each step. Most notably, how the state holding the output $\texttt{y}$ is updated after $\bm p_3$ is processed. }
    \label{fig:vec2vec}
\end{figure}

\iftagged{neurips}{We proceed with building LSRL programs that are universal in-context approximators: one for approximating continuous functions ($\CCvec$), and one for maps between token sequences ($\CCtok$).}
{Now that we are equipped with LSRL, we can proceed to building LSRL programs that are universal in-context approximators.
We will describe two programs: one for approximating continuous functions ($\CCvec$), and one for approximating maps between token sequences ($\CCtok$). Formally, we construct a model $g_\text{vec}$ of the Linear RNN architecture (\Cref{eq:linear_rnn}) such that $\HHvec(g_\text{vec})$ is dense in $\CCvec$ and a model $g_\text{tok}$ such that $\HHtok(g_\text{tok})$ is dense in $\CCtok$.}

\subsection{Approximating continuous functions in $\CCvec$}
\label{sec:approximating_cont_functions}
The idea behind the approximation for continuous functions is to discretise the domain into a grid and approximate the function as constant in each cell of the grid.
This technique is commonly used for showing universal approximation using the step activation function \citep{blum1991approximation,scarselli1998universal}. 
However, it is not obvious how to implement this approach in-context when information across input tokens can be only combined linearly. 
Consider a target function $f:[0,1]^{d_\text{in}}{\to}[0,1]^{d_\text{out}}$ and a discretization step $\delta$.
Our approach is to describe the value of $f$ in each of the discretization cells as a single prompt token.
For the cell with lower bounds $l_1, \ldots, l_{d_\text{in}}$ and their respective upper bounds $l_1\texttt{+}\delta, \smalldots, l_{d_\text{in}}\texttt{+}\delta$, the corresponding prompt token is a $(d_\text{in}\texttt{+}d_\text{out}\texttt{+}1)$-dimensional vector:
\begin{equation}
    \bm p = [ \delta, l_1, \ldots, l_{d_\text{in}}, \bm{\bar y}_1, \ldots \bm {\bar y}_{d_\text{out}} ]^\top,
    \label{eq:continous_prompt_def}
\end{equation}
where $\bm{\bar y}$ is the value of $f$ at the centre of that cell: $\bm{\bar y} = f(l_1\texttt{+}\nicefrac{\delta}{2},\smalldots, l_{d_\text{in}}\texttt{+}\nicefrac{\delta}{2})$.
Each prompt token describes the size of the cell (the discretisation step $\delta$), its starting lower bound, and the value of the target function at the centre of the cell.
Thus, ${\lceil \nicefrac{1}{\delta} \rceil}^{d_\text{in}}$ such tokens, one for each cell, are sufficient to describe the piece-wise constant approximation of $f$. 
A query $\bm q' \in [0,1]^{d_\text{in}}$ can fall in only one of the cells.
We pad it with zeros and encode it as the first input element: $\bm q = [\bm q'^\top, \bm 0_{d_\text{out}+1}^\top]^\top$, followed by the prompt.
Our program will extract and save $\bm q'$ to a state and then process the prompt tokens one at a time until it finds the one whose cell contains $\bm q'$.
The target function value for this cell will be added to an accumulator state.
If the current cell does not contain $\bm q'$, then 0 is instead added.%
Hence, the accumulator's final value corresponds to the value of $f$ at the centre of the cell containing $\bm q'$. 
The full LSRL program is provided in \Cref{lst:continous_approximation} and an illustration for $d_\text{in}=2, d_\text{out}=1, \delta=\nicefrac{1}{2}$ is shown in \Cref{fig:vec2vec}.
The prompt length required to approximate an $L$-Lipschitz function $f$ (w.r.t. the $\ell_2$ norm) to precision $\epsilon$ is $N=(\nicefrac{2\epsilon}{L\sqrt{d_\text{in}}})^{\texttt{-}d_\text{in}} = \mathcal O ( \epsilon^{\texttt{-}d_\text{in}})$ (see \Cref{sec:error_bound} for the proof).
Asymptotically, this is as good as one can hope without further assumptions on the target function.
This is also better than the best known result for the same problem for transformers: $\mathcal O (\epsilon^{\texttt{-}10\texttt{-}14d_\text{in}\texttt{-}4d_\text{in}^2})$ in \citealt{petrov2024universal}.

\untagged{neurips}
{The LSRL program in \Cref{lst:continous_approximation} also allows us to perform \emph{streaming} universal in-context approximation for free.
As the discretization step $\delta$ is not hard-coded in the model, we can first provide prompts at a coarse grid and then iteratively add prompts at increasingly finer grids, each providing a correction of the estimate of the previous one.
Thus, if the computation is interrupted, or a compute budget is reached, our model will still output an approximation of the target function.}

\begin{figure}
    \centering
    \includegraphics[width=0.95\textwidth]{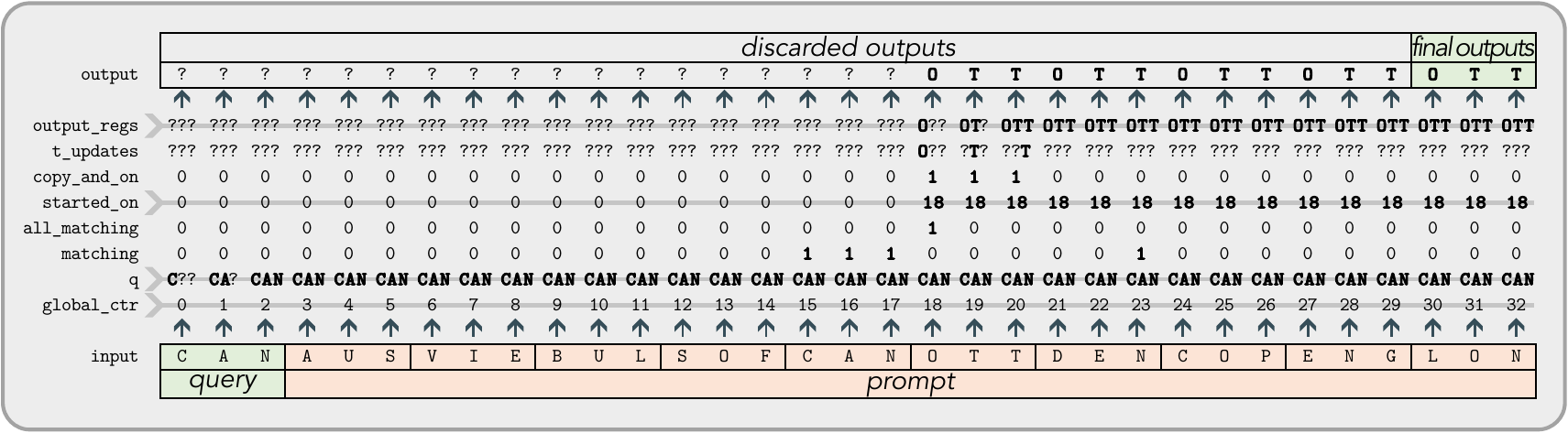}
    \caption{\textbf{Intuition behind the LSRL program for universal in-context approximation for discrete functions in \Cref{lst:tok2tok_approximation}.} Our keys and values have length $n{=}3$ and represent countries and capitals, e.g., \texttt{AUS}tria$\mapsto$\texttt{VIE}nna, \texttt{BUL}garia$\mapsto$\texttt{SOF}ia, and so on.
        The query is \texttt{CAN} for Canada and the final $n$ outputs are \texttt{OTT} (Ottawa). 
We show the values of some of the variables in \Cref{lst:tok2tok_approximation} at each step, with the \LinState variables being marked with arrows. For cleaner presentation we are tokenizing letters as \texttt{0}$\mapsto$\texttt{?},  \texttt{1}$\mapsto$\texttt{A}, \texttt{2}$\mapsto$\texttt{B}, etc. Vertical separators are for illustration purposes only.}
    \label{fig:tok2tok}
\end{figure}

\subsection{Approximating functions over token sequences in $\CCtok$}
\label{sec:approximating_discrete_functions}

\Cref{sec:approximating_cont_functions} focused on continuous functions but recurrent architectures are often used to model natural language whose domain is tokens.
Thus, we also look at modelling maps over a discrete domain.
Any function from $n$ tokens to $n$ tokens taking values in $\mathcal V=\{1,\dots,V\}$ can be represented as a dictionary whose keys and values are in $\mathcal{V}^n$.
Therefore, a simple way to represent this function in-context is to first provide the $n$ tokens corresponding to the query and then a sequence of $2n$ tokens corresponding to key and value pairs (see \Cref{fig:tok2tok} for an illustration of the setup).
The model stores the query in a state and processes the key-value pairs one by one by comparing the key (the first $n$ tokens) with the query.
If they match, then the value (the next $n$ tokens) is copied into a state that keeps it and repeatedly outputs it.
This continues until the end of the prompt, at which point the last $n$ outputted tokens will be the value corresponding to the key matching the query.
This is essentially a dictionary lookup.
However, as shown in \Cref{lst:tok2tok_approximation}, implementing dictionary lookup in a linear recurrent model is much less straightforward than executing \texttt{dict[key]} in a general-purpose programming language.

\Cref{lst:tok2tok_approximation} can appear daunting at first so we would like to clarify the non-trivial aspects.
First, we need to count how far we are into every set of $n$ or $2n$ tokens.
This can be done with $\Mod n$ and $\Mod 2n$ operations but implementing modulo for arbitrary large inputs is not possible with \ReLU MLPs \citep{ziyin2020neural}.
Therefore, we implement this with \LinState as \texttt{f\_modulo\_counter} which has a unit-length state that is rotated $\nicefrac{1}{n}$ or $\nicefrac{1}{2n}$ revolutions per iteration, with the angle corresponding to the modulo value (\Cref{sec:f_modulo_counter}).
Second, we need to do dynamic indexing to copy the $i$-th input in a subsequence to the $i$-th element of a state and vice-versa. 
Dynamic indexing, however, cannot be succinctly represented in a Linear RNN.
We work around this with temporary variables that are non-zero only at the $i$-th coordinates (see \Cref{line:tqs1,line:tqs2,line:qs1,line:qs2,line:outreg1,line:outreg2,line:outreg3,line:outreg4,line:out1,line:out2}).
Finally, in order to compare whether all $n$ elements in the query and the key match, we need to remember whether the previous $n$ pairs were matching.
As RNNs do not have attention, we implement this short-term memory buffer as a \LinState with a shift matrix (\Cref{line:buffer}).

\begin{lstfloat}[t]
\begin{lstlisting}[escapechar=@]
ForEach:
input = Input(dim=1)

# counter needed to know whether we are looking at the query or the prompt
const_1 = f_constant(input, 1)
global_ctr = LinState(input=const_1, A=[[1]], B=[[1]], init_state=[[-1]])

# counters mod[n] and mod[2n]
mod_n_ctr = f_modulo_counter(input, n)
mod_2n_ctr = f_modulo_counter(input, 2*n)

# which mode are we in (looking at the query, comparing query with key, or copying value to state)
is_prompt = f_larger(global_ctr, n-0.5)
is_compare_mode = f_larger(mod_2n_ctr, n-0.5)
is_copy_mode = f_and(is_prompt, f_not(is_compare_mode))
is_first_token_for_copy = f_and(is_copy_mode,f_smaller(mod_n_ctr, 0.5))

# update the state holding the query if this is one of the first n tokens
tq=f_ifelse(f_smaller(is_prompt, 0.5), t=input, f=input*0)
tqs=[f_ifelse(f_and(f_larger(mod_n_ctr,i-0.5), f_smaller(mod_n_ctr,i+0.5)),t=tq,f=tq*0) for i in 1..n]@\label{line:tqs1}@
q = LinState(input=Concat(tqs), A=eye(n), B=eye(n), init_state=zeros(n,1)) # query@\label{line:tqs2}@

# if we are in compare mode (looking at keys), check if this token matches the corresponding one in the query
qs=[f_ifelse(f_and(f_larger(mod_n_ctr,i-0.5),f_smaller(mod_n_ctr,i+0.5)),t=q[i],f=q[i]*0) for i in 1..n]@\label{line:qs1}@
cor_q_el = Linear(input=Concat(qs), A=ones(1,n), b=zeros(1,1))@\label{line:qs2}@
matching = f_and(f_and(f_larger(input, cor_q_el-0.5),f_smaller(input, cor_q_el+0.5)),is_compare_mode)

# keep a buffer of the last last n+1 match values, the +1 because we can only read the buffer after writing
buffer = LinState(input=matching, A=shift_matrix, B=[[0] for _ in 1..n], [[1]]), init_state=zeros(n+1,1))@\label{line:buffer}@
buffer_sum = Linear(input=buffer, A=[[1 for _ in 1..n], [0]], b=zeros(1, 1))
all_matching = f_larger(buffer_sum, n-0.5)

# if all are matching and it's the first token in the value part of the (key, value) pair, then mark this as the iterations when we start copying to state
matching_and_first_for_copy = f_and(all_matching, is_first_token_for_copy)
t_started_on_update = f_ifelse(matching_and_first_for_copy,t=global_ctr, f=global_ctr*0)
started_on = LinState(input=t_started_on_update, A=eye(1), B=eye(1), init_state=zeros(1, 1))

# copying to state for n iterations after started_on
copy_and_on = f_and(is_copy_mode, f_smaller(global_ctr, started_on+n))
mod_n_eq_i = [ f_and(f_larger(mod_n_ctr,i-0.5), f_smaller(mod_n_ctr,i+0.5)) for i in 1..n ]@\label{line:outreg1}@
t_updates_should_update = [f_and(copy_and_on, mod_n_eq_i[i]) for i in 1..n]@\label{line:outreg2}@
t_updates = [f_ifelse(f_larger(t_updates_should_update[i], 0.5), t=input, f=input*0) for i in 1..n]@\label{line:outreg3}@
output_regs = [LinState(input=update, A=eye(1), B=eye(1), init_state=zeros(1,1)) for update in t_updates]@\label{line:outreg4}@

# finally, read out the value from the corresponding output register in order to output from the model
t_outputs = [f_ifelse(f_larger(mod_n_eq_i[i], 0.5), t=output_regs[i], f=output_regs[i]*0) for i in 1..n]@\label{line:out1}@

return Linear(input=Concat(t_outputs), A=ones(1,n), b=zeros(1,1))@\label{line:out2}@
\end{lstlisting}
\caption{\textbf{LSRL program for universal in-context approximation of discrete functions.} The inputs are $\bm q_1,\smalldots,\bm q_n$ (the query tokens), followed by pairs of keys and values from the map we are approximating. The last $n$ outputs are the value corresponding to the key  matching the query. }
\label{lst:tok2tok_approximation}
\end{lstfloat}

\section{Stable Universal In-Context Approximation with Gated Linear RNNs}
\label{sec:ua_gated_linear_RNNs}

\vspace{-0.7em}
\paragraph{The ReLU-based conditional operator is not numerically stable.}
The LSRL programs in \Cref{lst:continous_approximation,lst:tok2tok_approximation} for  approximating functions in respectively $\CCvec$ and $\CCtok$ rely on the $\texttt{f\_ifelse}$ conditional assignment operator in \Cref{eq:ifelse_direct} in order to implement different behaviours depending on whether we are processing the query or specific parts of the prompt.
This operator is not numerically stable.
The first term in \Cref{eq:ifelse_direct} relies on $\texttt{cond}(\bm x)$ being exactly zero if the condition is not met.
In this way, multiplying it with $-\lambda$ would be 0 and $\texttt{f}(\bm x)$ would be returned.
However, if $\texttt{cond}(\bm x)$ is not identically 0 but has a small positive value, then $-\lambda \texttt{cond}(\bm x)$ can ``overpower'' $\texttt{f}(\bm x)$ resulting in the \ReLU output being 0.
In our experience, this is not a problem when processing inputs through the LSRL program step-by-step.
However, de-branching the DAG into a path graph ---which is necessary in order to uncover the equivalent Linear RNN--- appears to introduce such numerical instabilities which occasionally result in wrong outputs as conditional assignments will be 0 when they should not.
This problem is more prominent in \Cref{lst:tok2tok_approximation} which is longer (more debranching steps) and has more \texttt{f\_ifelse} operations: it gets most tokens wrong because of that instability (see \emph{Original, No noise} in \Cref{fig:ifelse_noise}).
To this end, we support LSRL with a symbolic backend (based on SymPy) that performs the debranching steps exactly.
Using it, both programs always produce the correct output.

This numerical instability highlights a critical practical limitations of the universal approximation results in \Cref{sec:ua_linear_RNNs}: if the models are not numerically stable, it is unlikely that they occur in practice by training models using gradient descent.
This section shows how to improve the numerical stability of \Cref{eq:ifelse_direct} and obtain more realistic recurrent models that are universal approximators in-context.

\begin{untaggedblock}{neurips}
\vspace{-0.7em}
\paragraph{Implementing \texttt{f\_ifelse} with MLPs.}
As LSRL allows us to express arbitrary MLPs and MLPs can approximate any continuous function, it is tempting to replace \Cref{eq:ifelse_direct} with a deep MLP model.
However, such implementation would also not be exact, which can cause problems when composing such logical operations.
For this reason, and for compactness of the resulting compiled model, we do not consider deeper implementations of \texttt{f\_ifelse}.
\end{untaggedblock}

\vspace{-0.7em}
\paragraph{Removing unnecessary terms in \Cref{eq:ifelse_direct}.}
\Cref{eq:ifelse_direct} has 4 separate \ReLU terms.
The first two handle the cases when $\texttt{t}(\bm x)$ and $\texttt{f}(\bm x)$ are positive and the second two when they are negative.
Therefore, if we know that one or both of these will always be non-negative, we can drop the corresponding terms.
\untagged{neurips}{This is especially useful for \Cref{lst:tok2tok_approximation}, approximating $\CCtok$, as it exclusively uses non-negative values (counters, mode switches and token values).}
Additionally, if $\texttt{f}(\bm x)$ is always $0$, then the first and third terms can be safely dropped.
Similarly, the second and fourth are unnecessary if $\texttt{f}(\bm x) \equiv 0$.
All $\texttt{f\_ifelse}$ in \Cref{lst:continous_approximation,lst:tok2tok_approximation} fall in this case and hence can be simplified.
We will refer to this $\texttt{f\_ifelse}$ implementation that is aware of the attainable values of $\texttt{t}(\bm x)$ and $\texttt{f}(\bm x)$ as \texttt{optimized}.
As it reduces the number of numerically unstable \ReLU operations in the model, we expect that it will improve the stability of the compiled models.
We experimented with adding various levels of noise to the non-zero model parameters, and, as the results in \Cref{fig:ifelse_noise} show, \texttt{optimized} is indeed more numerically robust than \texttt{original}.

\begin{figure}
    \centering
    \includegraphics[width=0.9\textwidth]{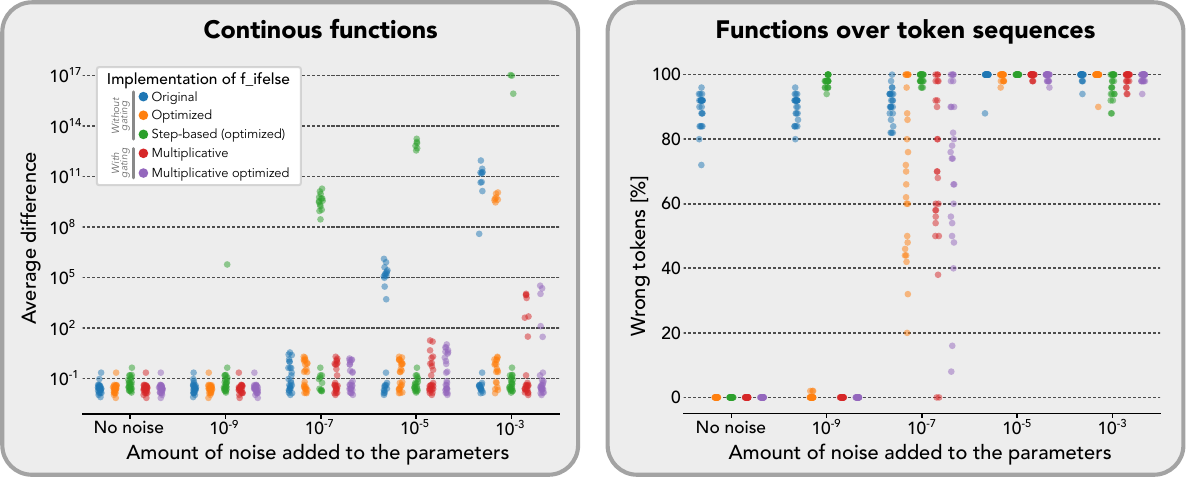}
    \caption{\textbf{Robustness of the various \texttt{f\_ifelse} implementations to model parameter noise.} We show how the performance of the two universal approximation programs in \Cref{lst:continous_approximation,lst:tok2tok_approximation} deteriorates as we add Gaussian noise of various magnitudes to the non-zero weights of the resulting compiled models. As expected, the original \texttt{f\_ifelse} implementation in \Cref{eq:ifelse_direct} exhibits numerical precision errors at the lowest noise magnitude.
    For the token sequence case, numerical precision errors are present in all samples even in the no-noise setting.
    Hence, the original \texttt{f\_ifelse} implementation is less numerically robust while the implementations with multiplicative gating are the most robust. For \Cref{lst:continous_approximation} (approximating $\CCvec$) we report the Euclidean distance between the target function value and the estimated one over 10 queries for 25 target functions. For \Cref{lst:tok2tok_approximation} we report the percentage of wrong token predictions over 5 queries for 25 dictionary maps. Lower values are better in both cases.}
    \label{fig:ifelse_noise}
\end{figure}

\vspace{-0.7em}
\paragraph{Step-based implementation.}
We can get rid of the input sensitivity of \Cref{eq:ifelse_direct} using \texttt{f\_step}:
\begin{equation}
    \resizebox{0.94\textwidth}{!}{$
    \begin{aligned}
    \texttt{f\_ifelse}[\texttt{cond}, \texttt{t}, \texttt{f}, \const{\lambda}](\bm x)
    &:=
    \ReLU(\texttt{-}\const{\lambda} \texttt{+} \lambda \texttt{f\_step}(\nicefrac{1}{2} \texttt{-} \texttt{cond}(\bm x)) \texttt{+} \texttt{f}(\bm x))
    + \ReLU(\texttt{-}\const{\lambda}\texttt{+} \lambda \texttt{f\_step}(\texttt{cond}(\bm x) \texttt{-}  \nicefrac{1}{2} ) \texttt{+} \texttt{t}(\bm x)) \\
    &\hspace{0.36em}- \ReLU(\texttt{-}\const{\lambda} \texttt{+} \lambda \texttt{f\_step}(\nicefrac{1}{2} \texttt{-} \texttt{cond}(\bm x)) \texttt{-} \texttt{f}(\bm x))
    - \ReLU(\texttt{-}\const{\lambda}\texttt{+} \lambda \texttt{f\_step}(\texttt{cond}(\bm x) \texttt{-}  \nicefrac{1}{2} ) \texttt{-} \texttt{t}(\bm x)).
    \end{aligned}
$}
    \label{eq:ifelse_step}
\end{equation}
We can also apply the optimisation strategy here. 
While this implementation is robust to noise in the input it appears to be more sensitive to parameter noise, as shown in \Cref{fig:ifelse_noise}.

\vspace{-0.7em}
\paragraph{Numerically stable $\texttt{f\_ifelse}$ with multiplicative gates.}
Removing the unused \ReLU terms in the original $\texttt{f\_ifelse}$ reduces the opportunities for numerical precision issues to creep in but does not solve the underlying problem.
The multiplicative gating present in the Linear Gated RNN (\Cref{eq:gated_linear_rnn}) and Gated RNN models (\Cref{eq:gated_rnn}) can help via implementing a numerically stable conditional operator:
\begin{equation}
    \texttt{f\_ifelse}[\texttt{cond}, \texttt{t}, \texttt{f}](\bm x)
        := \texttt{cond}(\bm x) \odot \texttt{t}(\bm x) 
        + 
        \texttt{f\_not}(\texttt{cond}(\bm x)) \odot \texttt{f}(\bm x),
        \label{eq:ifelse_multi}
\end{equation}
where the element-wise product is implemented in LSRL with \Concat and \Multi.
We will refer to the implementation of $\texttt{f\_ifelse}$ in \Cref{eq:ifelse_multi} as \texttt{multiplicative}.
Similarly to original implementation of $\texttt{f\_ifelse}$ in \Cref{eq:ifelse_direct}, we can drop the $\texttt{t}(\bm x)$ and $\texttt{f}(\bm x)$ term if they are equal to zero (\texttt{multiplicative optimized}).
If $\texttt{cond}(\bm x)$ is not exactly zero, $\texttt{cond}(\bm x) \odot \texttt{t}(\bm x)$ will result in a small error to the output but, in contrast to the original implementation, is not going to cause a discontinuity in the output of the operation. 
Therefore, \Cref{eq:ifelse_multi} should be more robust to numerical precision issues than \Cref{eq:ifelse_direct}.
\Cref{fig:ifelse_noise} shows that this is the case in practice with \Cref{lst:continous_approximation,lst:tok2tok_approximation} being more robust to parameter noise when using multiplicative gates compared to the \ReLU-based implementations.
Therefore, Linear Gated RNNs (\Cref{eq:gated_linear_rnn}) ---to which models with multiplicative gates can be compiled--- are more likely than Linear RNNs (\Cref{eq:linear_rnn}) to exhibit universal approximation properties in practice.

\begin{untaggedblock}{neurips}
Stability benefits of multiplicative gating in recurrent models has been previously shown in the context of deterministic finite-state automata \citep{omlin1996constructing}.
Beyond more stable conditional operators, multiplicative gating also results in strictly more expressive models than models with only element-wise nonlinearities \citep{jayakumar2020multiplicative}.
\end{untaggedblock}

\section{Universal In-context Approximation with Non-linear (Gated) RNNs}
\label{sec:ua_nonlinear_RNNs}
\Cref{sec:ua_linear_RNNs,sec:ua_gated_linear_RNNs} showed how universal approximation of continuous and token-to-token functions can be implemented in LSRL and compiled to respectively Linear RNNs and Linear Gated RNNs.
This section aims to address the situation with \emph{non-linear} state updates, that is, the cases of classic and gated RNNs (\Cref{eq:rnn,eq:gated_rnn}).
Concretely, we show how every \emph{linear} (Gated) RNN can be converted to a \emph{non-linear} (Gated) RNN.
\untagged{neurips}{Thus, we can compile any LSRL program (including \Cref{lst:continous_approximation,lst:tok2tok_approximation}) also to an RNN (if it has no \Multi operations) or a Gated RNN.\par}
The key idea is that the \ReLU applied to the state updates in the non-linear architectures is an identity operation if its inputs are positive.
Hence, we can split the states in positive and negative components, flip the sign of the negative component, pass them separately through the \ReLU ---which will act as an identity as all elements will be non-negative--- and then fuse the positive and negative components back together in the $\bm A$ matrix at the next time step\iftagged{neurips}{:}{Formally, we can convert a Linear RNN state update into a classic RNN state update as following:}
\begin{equation}
    \resizebox{0.85\textwidth}{!}{$
    \begin{aligned}
        \bm s_t &= \bm A \bm s_{t-1} + \bm B \bm x_t + \bm b \\
        \bm y_t   &= \phi(\bm s_t).
    \end{aligned}
    \hspace{1.0em}\equiv\hspace{1.0em}
    \begin{aligned}
        \begin{bmatrix} \bm s_t^+ \\ \bm s_t^- \end{bmatrix}
        &= \ReLU \left(
            \begin{bmatrix} \phantom{-}\bm A & -\bm A \\  -\bm A & \phantom{-}\bm A \end{bmatrix} 
            \begin{bmatrix} \bm s_t^+ \\ \bm s_t^- \end{bmatrix}
            +
            \begin{bmatrix} \phantom{-}\bm B \\ -\bm B \end{bmatrix}
            \bm x_t + \begin{bmatrix} \phantom{-}\bm b \\ -\bm b \end{bmatrix} \right) \\
        \bm y_t   &= \phi \left( \begin{bmatrix} \bm I & -\bm I \end{bmatrix} \begin{bmatrix} \bm s_t^+ \\ \bm s_t^- \end{bmatrix} \right).
    \end{aligned}$}
    \label{eq:linear_to_nonlinear}
\end{equation}

\begin{untaggedblock}{neurips}
Another way to look at this is by recognizing that an RNN is equivalent to a Linear RNN with the exact same weights if the states are always non-negative.
Hence, all we need is a trick to ensure the states are non-negative.
This approach works just as well for the Gated RNNs as the gating and the state updates are independent from one another.
\end{untaggedblock}

Using \Cref{eq:linear_to_nonlinear} we can compile any LSRL program to an RNN (\Cref{eq:rnn}) or a Gated RNN (\Cref{eq:gated_rnn}).
This includes \Cref{lst:continous_approximation,lst:tok2tok_approximation}.
Hence, RNNs and Gated RNNs can be universal in-context approximators for continuous and token-to-token functions.
As any Gated RNN can be represented as a GRU model (\Cref{sec:GRU}) or an LSTM (\Cref{sec:lstms}), these models are too universal in-context approximators.
\untagged{neurips}{The same numerical stability issues discussed in \Cref{sec:ua_gated_linear_RNNs} apply here and as a result, universal approximation capabilities are probably more likely to occur in Gated RNNs than in RNNs.}

\section{Discussion and Conclusions}
\label{sec:discussion}
We developed LSRL: a programming language for specifying programs expressible with recurrent neural architectures.
We then used LSRL to show that various architectures ---from the humble RNN to the state-of-the-art Linear Gated RNNs--- can all be universal approximators \emph{in-context}.
\untagged{neurips}{That is, there exist fixed models with these architectures which can be prompted to act as any token-to-token function or approximate any continuous function to an arbitrary precision. These results do not assume  infinite precision or exponential hidden state sizes. The hidden state sizes of our constructions are also independent of the target precision (in the continuous setting) or the key-value dictionary size (in the discrete setting). }

\begin{untaggedblock}{neurips}
\vspace{-0.7em}
\paragraph{Comparisson with the transformer architecture.}
Contemporary Linear RNNs attempt to challenge the dominant role of the transformer architecture.
At the same time, our understanding of their in-context abilities is significantly lacking behind that of the transfomer.
This work makes an important contribution to this problem: we showed that Linear SSMs are not only universal in-context approximators but are potentially require shorter prompts than transformers ($\mathcal O ( \epsilon^{-d_\text{in}})$ vs $\mathcal O (\epsilon^{-10-14d_\text{in}-4d_\text{in}^2})$ from \citealt{petrov2024universal}).
That approach also relies on the Kolmogorov-Arnold representation theorem \citep{kolmogorov1957representation} which is notoriously unlikely to be useful in practice \citep{girosi1989representation}.
Our constructions are much simpler, especially in the token-to-token case.

\vspace{-0.7em}
\paragraph{Ability of models to implement algorithms.}
There has been a lot of attention on evaluating how well models can execute various algorithms \citep{giannou23looped,la2024code,sanford2024transformers}. 
These abilities are fundamentally limited to how many basic operations a model can do in a single forward pass.
Our results indicate that this might be architecture dependent and that multiplicative gating might be much more efficient for implementing logic operations at the core of many algorithms.
Hence, it might be possible that more complex logic programs could be expressed with the same number of parameters if one uses an architecture with multiplicative gating.

\vspace{-0.7em}
\paragraph{Safety and security implications.}
If a model can be prompted to approximate any function, then preventing it from exhibiting undesirable behaviours (i.e., alignment) might be fundamentally impossible.
On the flip side, recently, methods for improving the safety of transformers using  interpretablity approaches have been proposed \citep{conmy2023towards,geiger2024finding}.
The success of  interpretability techniques is difficult to assess though.
To assist with that, benchmarks of models with known behaviours can be developed.
RASP has already been used to evaluate \emph{transformer} interpretablity methods \citep{friedman2024learning,zhou2023algorithms}.
However, interpretability tools for fully recurrent are significantly lagging behind the ones for transformers.
Therefore, we hope that LSRL can be helpful for designing interpretability benchmarks for fully recurrent models, similarly to how RASP has contributed to understanding  transformer models.
\end{untaggedblock}

\begin{taggedblock}{neurips}
\vspace{-0.7em}
\paragraph{Safety and security implications.}
If a model can be prompted to approximate any function, then preventing it from exhibiting undesirable behaviours (i.e., alignment) might be impossible.
Therefore, it is important to further study the safety and security implications of these properties.
\end{taggedblock}

\vspace{-0.7em}
\paragraph{Limitations.}
In this work we provide \emph{constructive existence results}: that is, we show that there can exist models with various recurrent architectures that are universal in-context approximators.
However, the present theory is not sufficient to analyse whether \emph{a given model} has this property.
That is a much more difficult question that would require a very different approach.
We also assume no restrictions on the $\bm A$ matrix in the state update equations.
However, many state-of-the-art models impose structural constraints on $\bm A$ (e.g., it being diagonal) for the sake of fast training and inference \citep{gu2020hippo,gu2021efficiently,gupta2022diagonal}.
It is not directly obvious whether such structural restrictions would affect the universal in-context approximation abilities of these architectures.
In practice, however, the compiled matrices are very sparse and often diagonal.
Therefore, it is highly likely that our results translate to models with structural restrictions.

\section*{Acknowledgements}
We would like to thank Simon Schug for pointing us to relevant works on fully recurrent models.
We are also extremely grateful to Juuso Haavisto for his insight on building compilers.
This work is supported by a UKRI grant Turing AI Fellowship (EP/W002981/1) and the EPSRC Centre for Doctoral Training in Autonomous Intelligent Machines and Systems (EP/S024050/1).

\bibliography{bibliography.bib}

\begin{thebibliography}{71}
\expandafter\ifx\csname natexlab\endcsname\relax\def\natexlab#1{#1}\fi

\bibitem[{Ahn et~al.(2023)Ahn, Cheng, Daneshmand, and
  Sra}]{ahn2023transformers}
Kwangjun Ahn, Xiang Cheng, Hadi Daneshmand, and Suvrit Sra. 2023.
\newblock \href
  {https://proceedings.neurips.cc/paper_files/paper/2023/hash/8ed3d610ea4b68e7afb30ea7d01422c6-Abstract-Conference.html}
  {Transformers learn to implement preconditioned gradient descent for
  in-context learning}.
\newblock In \emph{Advances in Neural Information Processing Systems}.

\bibitem[{Aky{\"u}rek et~al.(2022)Aky{\"u}rek, Schuurmans, Andreas, Ma, and
  Zhou}]{akyurek2022learning}
Ekin Aky{\"u}rek, Dale Schuurmans, Jacob Andreas, Tengyu Ma, and Denny Zhou.
  2022.
\newblock \href {https://openreview.net/forum?id=0g0X4H8yN4I} {What learning
  algorithm is in-context learning? {Investigations} with linear models}.
\newblock In \emph{The Eleventh International Conference on Learning
  Representations}.

\bibitem[{Aky{\"u}rek et~al.(2024)Aky{\"u}rek, Wang, Kim, and
  Andreas}]{akyurek2024context}
Ekin Aky{\"u}rek, Bailin Wang, Yoon Kim, and Jacob Andreas. 2024.
\newblock \href {https://arxiv.org/abs/2401.12973} {In-context language
  learning: Arhitectures and algorithms}.
\newblock \emph{arXiv preprint arXiv:2401.12973}.

\bibitem[{Amari(1972)}]{amari1992rnns}
Shun-ichi Amari. 1972.
\newblock \href {https://doi.org/10.1109/T-C.1972.223477} {Learning patterns
  and pattern sequences by self-organizing nets of threshold elements}.
\newblock \emph{IEEE Transactions on Computers}, C-21(11):1197--1206.

\bibitem[{Ba et~al.(2016)Ba, Kiros, and Hinton}]{ba2016layer}
Jimmy~Lei Ba, Jamie~Ryan Kiros, and Geoffrey~E Hinton. 2016.
\newblock \href {https://arxiv.org/abs/1607.06450} {Layer normalization}.
\newblock \emph{arXiv preprint arXiv:1607.06450}.

\bibitem[{Bahdanau et~al.(2015)Bahdanau, Cho, and Bengio}]{bahdanau2014neural}
Dzmitry Bahdanau, Kyunghyun Cho, and Yoshua Bengio. 2015.
\newblock \href {https://arxiv.org/abs/1409.0473} {Neural machine translation
  by jointly learning to align and translate}.
\newblock In \emph{International Conference on Learning Representations}.

\bibitem[{Bai et~al.(2023)Bai, Chen, Wang, Xiong, and
  Mei}]{bai2024transformers}
Yu~Bai, Fan Chen, Huan Wang, Caiming Xiong, and Song Mei. 2023.
\newblock \href
  {https://proceedings.neurips.cc/paper_files/paper/2023/hash/b2e63e36c57e153b9015fece2352a9f9-Abstract-Conference.html}
  {Transformers as statisticians: Provable in-context learning with in-context
  algorithm selection}.
\newblock In \emph{Advances in neural information processing systems}.

\bibitem[{Barron(1993)}]{barron1993universal}
Andrew~R Barron. 1993.
\newblock \href {https://doi.org/10.1109/18.256500} {Universal approximation
  bounds for superpositions of a sigmoidal function}.
\newblock \emph{IEEE Transactions on Information Theory}, 39(3):930--945.

\bibitem[{Beltagy et~al.(2020)Beltagy, Peters, and
  Cohan}]{beltagy2020longformer}
Iz~Beltagy, Matthew~E Peters, and Arman Cohan. 2020.
\newblock \href {https://arxiv.org/abs/2004.05150} {Longformer: The
  long-document transformer}.
\newblock \emph{arXiv preprint arXiv:2004.05150}.

\bibitem[{Bengio et~al.(1994)Bengio, Simard, and Frasconi}]{bengio1994learning}
Yoshua Bengio, Patrice Simard, and Paolo Frasconi. 1994.
\newblock \href {https://ieeexplore.ieee.org/document/279181} {Learning
  long-term dependencies with gradient descent is difficult}.
\newblock \emph{IEEE Transactions on Neural Networks}, 5(2):157--166.

\bibitem[{Blum and Li(1991)}]{blum1991approximation}
Edward~K Blum and Leong~Kwan Li. 1991.
\newblock Approximation theory and feedforward networks.
\newblock \emph{Neural networks}, 4(4):511--515.

\bibitem[{Botev et~al.(2024)Botev, De, Smith, Fernando, Muraru, Haroun,
  Berrada, Pascanu, Sessa, Dadashi et~al.}]{botev2024recurrentgemma}
Aleksandar Botev, Soham De, Samuel~L Smith, Anushan Fernando, George-Cristian
  Muraru, Ruba Haroun, Leonard Berrada, Razvan Pascanu, Pier~Giuseppe Sessa,
  Robert Dadashi, et~al. 2024.
\newblock \href {https://arxiv.org/abs/2404.07839} {{RecurrentGemma}: Moving
  past transformers for efficient open language models}.
\newblock \emph{arXiv preprint arXiv:2404.07839}.

\bibitem[{Boyd and Chua(1985)}]{boyd1985fading}
Stephen Boyd and Leon Chua. 1985.
\newblock Fading memory and the problem of approximating nonlinear operators
  with {Volterra} series.
\newblock \emph{IEEE Transactions on Circuits and Systems}, 32(11):1150--1161.

\bibitem[{Brown et~al.(2020)Brown, Mann, Ryder, Subbiah, Kaplan, Dhariwal,
  Neelakantan, Shyam, Sastry, Askell et~al.}]{brown2020language}
Tom Brown, Benjamin Mann, Nick Ryder, Melanie Subbiah, Jared~D Kaplan, Prafulla
  Dhariwal, Arvind Neelakantan, Pranav Shyam, Girish Sastry, Amanda Askell,
  et~al. 2020.
\newblock \href
  {https://papers.nips.cc/paper/2020/hash/1457c0d6bfcb4967418bfb8ac142f64a-Abstract.html}
  {Language models are few-shot learners}.
\newblock In \emph{Advances in Neural Information Processing Systems}.

\bibitem[{Cho et~al.(2014)Cho, van Merri{\"e}nboer, Gulcehre, Bahdanau,
  Bougares, Schwenk, and Bengio}]{cho2014learning}
Kyunghyun Cho, Bart van Merri{\"e}nboer, Caglar Gulcehre, Dzmitry Bahdanau,
  Fethi Bougares, Holger Schwenk, and Yoshua Bengio. 2014.
\newblock \href {https://aclanthology.org/D14-1179} {Learning phrase
  representations using {RNN} encoder{--}decoder for statistical machine
  translation}.
\newblock In \emph{Proceedings of the 2014 Conference on Empirical Methods in
  Natural Language Processing ({EMNLP})}.

\bibitem[{Coda-Forno et~al.(2023)Coda-Forno, Binz, Akata, Botvinick, Wang, and
  Schulz}]{coda2023meta}
Julian Coda-Forno, Marcel Binz, Zeynep Akata, Matt Botvinick, Jane Wang, and
  Eric Schulz. 2023.
\newblock \href
  {https://proceedings.neurips.cc/paper_files/paper/2023/hash/cda04d7ea67ea1376bf8c6962d8541e0-Abstract-Conference.html}
  {Meta-in-context learning in large language models}.
\newblock In \emph{Advances in Neural Information Processing Systems}, pages
  65189--65201.

\bibitem[{Conmy et~al.(2023)Conmy, Mavor-Parker, Lynch, Heimersheim, and
  Garriga-Alonso}]{conmy2023towards}
Arthur Conmy, Augustine Mavor-Parker, Aengus Lynch, Stefan Heimersheim, and
  Adri{\`a} Garriga-Alonso. 2023.
\newblock \href
  {https://proceedings.neurips.cc/paper_files/paper/2023/hash/34e1dbe95d34d7ebaf99b9bcaeb5b2be-Abstract-Conference.html}
  {Towards automated circuit discovery for mechanistic interpretability}.
\newblock \emph{Advances in Neural Information Processing Systems}.

\bibitem[{Cybenko(1989)}]{cybenko1989approximation}
George Cybenko. 1989.
\newblock \href {https://doi.org/10.1007/BF02551274} {Approximation by
  superpositions of a sigmoidal function}.
\newblock \emph{Mathematics of control, signals and systems}, 2(4):303--314.

\bibitem[{Dai et~al.(2023)Dai, Sun, Dong, Hao, Ma, Sui, and Wei}]{dai2023gpt}
Damai Dai, Yutao Sun, Li~Dong, Yaru Hao, Shuming Ma, Zhifang Sui, and Furu Wei.
  2023.
\newblock \href {https://aclanthology.org/2023.findings-acl.247} {Why can {GPT}
  learn in-context? {Language} models secretly perform gradient descent as
  meta-optimizers}.
\newblock In \emph{Findings of the Association for Computational Linguistics:
  ACL 2023}.

\bibitem[{De et~al.(2024)De, Smith, Fernando, Botev, Cristian-Muraru, Gu,
  Haroun, Berrada, Chen, Srinivasan et~al.}]{de2024griffin}
Soham De, Samuel~L Smith, Anushan Fernando, Aleksandar Botev, George
  Cristian-Muraru, Albert Gu, Ruba Haroun, Leonard Berrada, Yutian Chen,
  Srivatsan Srinivasan, et~al. 2024.
\newblock \href {https://arxiv.org/abs/2402.19427} {Griffin: Mixing gated
  linear recurrences with local attention for efficient language models}.
\newblock \emph{arXiv preprint arXiv:2402.19427}.

\bibitem[{Friedman et~al.(2023)Friedman, Wettig, and
  Chen}]{friedman2024learning}
Dan Friedman, Alexander Wettig, and Danqi Chen. 2023.
\newblock \href
  {https://proceedings.neurips.cc/paper_files/paper/2023/hash/995f693b73050f90977ed2828202645c-Abstract-Conference.html}
  {Learning transformer programs}.
\newblock In \emph{Advances in Neural Information Processing Systems}.

\bibitem[{Fu et~al.(2023{\natexlab{a}})Fu, Dao, Saab, Thomas, Rudra, and
  Re}]{fu2023hungry}
Daniel~Y Fu, Tri Dao, Khaled~Kamal Saab, Armin~W Thomas, Atri Rudra, and
  Christopher Re. 2023{\natexlab{a}}.
\newblock \href {https://openreview.net/forum?id=COZDy0WYGg} {{Hungry Hungry
  Hippos}: {Towards} language modeling with state space models}.
\newblock In \emph{International Conference on Learning Representations}.

\bibitem[{Fu et~al.(2023{\natexlab{b}})Fu, Chen, Jia, and
  Sharan}]{fu2023transformers}
Deqing Fu, Tian-Qi Chen, Robin Jia, and Vatsal Sharan. 2023{\natexlab{b}}.
\newblock \href {https://arxiv.org/abs/2310.17086} {Transformers learn
  higher-order optimization methods for in-context learning: A study with
  linear models}.
\newblock \emph{arXiv preprint arXiv:2310.17086}.

\bibitem[{Garg et~al.(2022)Garg, Tsipras, Liang, and Valiant}]{garg2022can}
Shivam Garg, Dimitris Tsipras, Percy~S Liang, and Gregory Valiant. 2022.
\newblock \href
  {https://proceedings.neurips.cc/paper_files/paper/2022/hash/c529dba08a146ea8d6cf715ae8930cbe-Abstract-Conference.html}
  {What can transformers learn in-context? {A} case study of simple function
  classes}.
\newblock In \emph{Advances in Neural Information Processing Systems}.

\bibitem[{Geiger et~al.(2024)Geiger, Wu, Potts, Icard, and
  Goodman}]{geiger2024finding}
Atticus Geiger, Zhengxuan Wu, Christopher Potts, Thomas Icard, and Noah
  Goodman. 2024.
\newblock Finding alignments between interpretable causal variables and
  distributed neural representations.
\newblock In \emph{Causal Learning and Reasoning}.

\bibitem[{Gers et~al.(2000)Gers, Schmidhuber, and Cummins}]{gers2000learning}
Felix~A Gers, J{\"u}rgen Schmidhuber, and Fred Cummins. 2000.
\newblock Learning to forget: Continual prediction with lstm.
\newblock \emph{Neural computation}, 12(10):2451--2471.

\bibitem[{Giannou et~al.(2023)Giannou, Rajput, Sohn, Lee, Lee, and
  Papailiopoulos}]{giannou23looped}
Angeliki Giannou, Shashank Rajput, Jy-Yong Sohn, Kangwook Lee, Jason~D. Lee,
  and Dimitris Papailiopoulos. 2023.
\newblock \href {https://proceedings.mlr.press/v202/giannou23a.html} {Looped
  transformers as programmable computers}.
\newblock In \emph{International Conference on Machine Learning}.

\bibitem[{Girosi and Poggio(1989)}]{girosi1989representation}
Federico Girosi and Tomaso Poggio. 1989.
\newblock \href {https://doi.org/10.1162/neco.1989.1.4.465} {Representation
  properties of networks: {Kolmogorov's} theorem is irrelevant}.
\newblock \emph{Neural Computation}, 1(4):465--469.

\bibitem[{Gu and Dao(2023)}]{gu2023mamba}
Albert Gu and Tri Dao. 2023.
\newblock \href {https://arxiv.org/abs/2312.00752} {Mamba: Linear-time sequence
  modeling with selective state spaces}.
\newblock \emph{arXiv preprint arXiv:2312.00752}.

\bibitem[{Gu et~al.(2020)Gu, Dao, Ermon, Rudra, and R{\'e}}]{gu2020hippo}
Albert Gu, Tri Dao, Stefano Ermon, Atri Rudra, and Christopher R{\'e}. 2020.
\newblock \href
  {https://proceedings.neurips.cc/paper/2020/hash/102f0bb6efb3a6128a3c750dd16729be-Abstract.html}
  {{HiPPO}: {Recurrent} memory with optimal polynomial projections}.
\newblock In \emph{Advances in Neural Information Processing Systems}.

\bibitem[{Gu et~al.(2021)Gu, Goel, and Re}]{gu2021efficiently}
Albert Gu, Karan Goel, and Christopher Re. 2021.
\newblock \href {https://openreview.net/forum?id=uYLFoz1vlAC} {Efficiently
  modeling long sequences with structured state spaces}.
\newblock In \emph{International Conference on Learning Representations}.

\bibitem[{Gupta et~al.(2022)Gupta, Gu, and Berant}]{gupta2022diagonal}
Ankit Gupta, Albert Gu, and Jonathan Berant. 2022.
\newblock \href
  {https://proceedings.neurips.cc/paper_files/paper/2022/hash/9156b0f6dfa9bbd18c79cc459ef5d61c-Abstract-Conference.html}
  {Diagonal state spaces are as effective as structured state spaces}.
\newblock In \emph{Advances in Neural Information Processing Systems}.

\bibitem[{Han et~al.(2023)Han, Wang, Zhao, and Ji}]{han2023context}
Chi Han, Ziqi Wang, Han Zhao, and Heng Ji. 2023.
\newblock \href {https://arxiv.org/abs/2305.12766} {In-context learning of
  large language models explained as kernel regression}.
\newblock \emph{arXiv preprint arXiv:2305.12766}.

\bibitem[{He et~al.(2016)He, Zhang, Ren, and Sun}]{he2016deep}
Kaiming He, Xiangyu Zhang, Shaoqing Ren, and Jian Sun. 2016.
\newblock \href
  {https://openaccess.thecvf.com/content_cvpr_2016/html/He_Deep_Residual_Learning_CVPR_2016_paper.html}
  {Deep residual learning for image recognition}.
\newblock In \emph{Proceedings of the IEEE Conference on Computer Vision and
  pattern Recognition}.

\bibitem[{Hochreiter and Schmidhuber(1997)}]{hochreiter1997long}
Sepp Hochreiter and J{\"u}rgen Schmidhuber. 1997.
\newblock \href {https://ieeexplore.ieee.org/abstract/document/6795963} {Long
  short-term memory}.
\newblock \emph{Neural Computation}, 9(8):1735--1780.

\bibitem[{Hornik et~al.(1989)Hornik, Stinchcombe, and
  White}]{hornik1989multilayer}
Kurt Hornik, Maxwell Stinchcombe, and Halbert White. 1989.
\newblock Multilayer feedforward networks are universal approximators.
\newblock \emph{Neural networks}, 2(5):359--366.

\bibitem[{Jayakumar et~al.(2020)Jayakumar, Czarnecki, Menick, Schwarz, Rae,
  Osindero, Teh, Harley, and Pascanu}]{jayakumar2020multiplicative}
Siddhant~M Jayakumar, Wojciech~M Czarnecki, Jacob Menick, Jonathan Schwarz,
  Jack Rae, Simon Osindero, Yee~Whye Teh, Tim Harley, and Razvan Pascanu. 2020.
\newblock \href {https://openreview.net/forum?id=rylnK6VtDH} {Multiplicative
  interactions and where to find them}.
\newblock In \emph{International Conference on Learning Representations}.

\bibitem[{Kolmogorov(1957)}]{kolmogorov1957representation}
Andrei~Nikolaevich Kolmogorov. 1957.
\newblock On the representation of continuous functions of many variables by
  superposition of continuous functions of one variable and addition.
\newblock In \emph{Doklady Akademii Nauk}, volume 114, pages 953--956. Russian
  Academy of Sciences.

\bibitem[{{La Malfa} et~al.(2023){La Malfa}, Petrov, Frieder, Weinhuber,
  Burnell, Nazar, Cohn, Shadbolt, and Wooldridge}]{lamalfa2023language}
Emanuele {La Malfa}, Aleksandar Petrov, Simon Frieder, Christoph Weinhuber,
  Ryan Burnell, Raza Nazar, Anthony~G. Cohn, Nigel Shadbolt, and Michael
  Wooldridge. 2023.
\newblock \href {https://arxiv.org/abs/2309.16573} {{Language Models as a
  Service}: Overview of a new paradigm and its challenges}.
\newblock \emph{arXiv preprint arXiv:2309.16573}.

\bibitem[{La~Malfa et~al.(2024)La~Malfa, Weinhuber, Torre, Lin, Cohn, Shadbolt,
  and Wooldridge}]{la2024code}
Emanuele La~Malfa, Christoph Weinhuber, Orazio Torre, Fangru Lin, Anthony Cohn,
  Nigel Shadbolt, and Michael Wooldridge. 2024.
\newblock \href {https://arxiv.org/abs/2401.09074} {Code simulation challenges
  for large language models}.
\newblock \emph{arXiv preprint arXiv:2401.09074}.

\bibitem[{Lee et~al.(2024)Lee, Jiang, and Berg-Kirkpatrick}]{lee2023exploring}
Ivan Lee, Nan Jiang, and Taylor Berg-Kirkpatrick. 2024.
\newblock \href {https://openreview.net/forum?id=Qwq4cpLtoX} {Exploring the
  relationship between model architecture and in-context learning ability}.
\newblock In \emph{International Conference on Learning Representations}.

\bibitem[{Li et~al.(2023)Li, Ildiz, Papailiopoulos, and
  Oymak}]{li2023transformers}
Yingcong Li, Muhammed~Emrullah Ildiz, Dimitris Papailiopoulos, and Samet Oymak.
  2023.
\newblock \href {https://proceedings.mlr.press/v202/li23l.html} {Transformers
  as algorithms: Generalization and stability in in-context learning}.
\newblock In \emph{International Conference on Machine Learning}.

\bibitem[{Lindner et~al.(2023)Lindner, Kram{\'a}r, Farquhar, Rahtz, McGrath,
  and Mikulik}]{lindner2024tracr}
David Lindner, J{\'a}nos Kram{\'a}r, Sebastian Farquhar, Matthew Rahtz, Tom
  McGrath, and Vladimir Mikulik. 2023.
\newblock \href
  {https://proceedings.neurips.cc/paper_files/paper/2023/hash/771155abaae744e08576f1f3b4b7ac0d-Abstract-Conference.html}
  {Tracr: {Compiled} transformers as a laboratory for interpretability}.
\newblock In \emph{Advances in Neural Information Processing Systems}.

\bibitem[{Liu et~al.(2023)Liu, Yuan, Fu, Jiang, Hayashi, and
  Neubig}]{liu2023pretrain}
Pengfei Liu, Weizhe Yuan, Jinlan Fu, Zhengbao Jiang, Hiroaki Hayashi, and
  Graham Neubig. 2023.
\newblock \href {https://dl.acm.org/doi/full/10.1145/3560815} {Pre-train,
  prompt, and predict: A systematic survey of prompting methods in natural
  language processing}.
\newblock \emph{ACM Computing Surveys}.

\bibitem[{Liu and Chilton(2022)}]{liu2022design}
Vivian Liu and Lydia~B Chilton. 2022.
\newblock \href {https://doi.org/10.1145/3491102.3501825} {Design guidelines
  for prompt engineering text-to-image generative models}.
\newblock In \emph{Proceedings of the 2022 CHI Conference on Human Factors in
  Computing Systems}.

\bibitem[{Meurer et~al.(2017)Meurer, Smith, Paprocki, \v{C}ert\'{i}k,
  Kirpichev, Rocklin, Kumar, Ivanov, Moore, Singh, Rathnayake, Vig, Granger,
  Muller, Bonazzi, Gupta, Vats, Johansson, Pedregosa, Curry, Terrel,
  Rou\v{c}ka, Saboo, Fernando, Kulal, Cimrman, and Scopatz}]{SymPy}
Aaron Meurer, Christopher~P. Smith, Mateusz Paprocki, Ond\v{r}ej
  \v{C}ert\'{i}k, Sergey~B. Kirpichev, Matthew Rocklin, AMiT Kumar, Sergiu
  Ivanov, Jason~K. Moore, Sartaj Singh, Thilina Rathnayake, Sean Vig, Brian~E.
  Granger, Richard~P. Muller, Francesco Bonazzi, Harsh Gupta, Shivam Vats,
  Fredrik Johansson, Fabian Pedregosa, Matthew~J. Curry, Andy~R. Terrel,
  \v{S}t\v{e}p\'{a}n Rou\v{c}ka, Ashutosh Saboo, Isuru Fernando, Sumith Kulal,
  Robert Cimrman, and Anthony Scopatz. 2017.
\newblock \href {https://doi.org/10.7717/peerj-cs.103} {{SymPy}: Symbolic
  computing in {Python}}.
\newblock \emph{PeerJ Computer Science}, 3.

\bibitem[{Omlin and Giles(1996)}]{omlin1996constructing}
Christian~W Omlin and C~Lee Giles. 1996.
\newblock \href {https://dl.acm.org/doi/abs/10.1145/235809.235811}
  {Constructing deterministic finite-state automata in recurrent neural
  networks}.
\newblock \emph{Journal of the ACM (JACM)}, 43(6):937--972.

\bibitem[{Orvieto et~al.(2023)Orvieto, Smith, Gu, Fernando, Gulcehre, Pascanu,
  and De}]{orvieto2023resurrecting}
Antonio Orvieto, Samuel~L Smith, Albert Gu, Anushan Fernando, Caglar Gulcehre,
  Razvan Pascanu, and Soham De. 2023.
\newblock \href {https://arxiv.org/abs/2303.06349} {Resurrecting recurrent
  neural networks for long sequences}.
\newblock In \emph{International Conference on Machine Learning}.

\bibitem[{Petrov et~al.(2024{\natexlab{a}})Petrov, Torr, and
  Bibi}]{petrov2024universal}
Aleksandar Petrov, Philip~HS Torr, and Adel Bibi. 2024{\natexlab{a}}.
\newblock \href {https://arxiv.org/abs/2402.14753} {Prompting a pretrained
  transformer can be a universal approximator}.
\newblock In \emph{International Conference on Machine Learning}.

\bibitem[{Petrov et~al.(2024{\natexlab{b}})Petrov, Torr, and
  Bibi}]{petrov2023prompting}
Aleksandar Petrov, Philip~HS Torr, and Adel Bibi. 2024{\natexlab{b}}.
\newblock \href {https://arxiv.org/abs/2310.19698} {When do prompting and
  prefix-tuning work? {A} theory of capabilities and limitations}.
\newblock In \emph{International Conference on Learning Representations}.

\bibitem[{Sahoo et~al.(2024)Sahoo, Singh, Saha, Jain, Mondal, and
  Chadha}]{sahoo2024systematic}
Pranab Sahoo, Ayush~Kumar Singh, Sriparna Saha, Vinija Jain, Samrat Mondal, and
  Aman Chadha. 2024.
\newblock \href {https://arxiv.org/abs/2402.07927} {A systematic survey of
  prompt engineering in large language models: Techniques and applications}.
\newblock \emph{arXiv preprint arXiv:2402.07927}.

\bibitem[{Sanford et~al.(2024)Sanford, Hsu, and
  Telgarsky}]{sanford2024transformers}
Clayton Sanford, Daniel Hsu, and Matus Telgarsky. 2024.
\newblock \href {https://arxiv.org/abs/2402.09268} {Transformers, parallel
  computation, and logarithmic depth}.
\newblock \emph{arXiv preprint arXiv:2402.09268}.

\bibitem[{Scarselli and Tsoi(1998)}]{scarselli1998universal}
Franco Scarselli and Ah~Chung Tsoi. 1998.
\newblock Universal approximation using feedforward neural networks: A survey
  of some existing methods, and some new results.
\newblock \emph{Neural networks}, 11(1):15--37.

\bibitem[{Sch{\"a}fer and Zimmermann(2006)}]{schafer2006recurrent}
Anton~Maximilian Sch{\"a}fer and Hans~Georg Zimmermann. 2006.
\newblock Recurrent neural networks are universal approximators.
\newblock In \emph{Artificial Neural Networks--ICANN 2006: 16th International
  Conference, Athens, Greece, September 10-14, 2006. Proceedings, Part I 16},
  pages 632--640. Springer.

\bibitem[{Shazeer(2019)}]{shazeer2019fast}
Noam Shazeer. 2019.
\newblock \href {https://arxiv.org/abs/1911.02150} {Fast transformer decoding:
  One write-head is all you need}.
\newblock \emph{arXiv preprint arXiv:1911.02150}.

\bibitem[{Su et~al.(2024)Su, Ahmed, Lu, Pan, Bo, and Liu}]{su2024roformer}
Jianlin Su, Murtadha Ahmed, Yu~Lu, Shengfeng Pan, Wen Bo, and Yunfeng Liu.
  2024.
\newblock \href {https://arxiv.org/abs/2104.09864} {Roformer: {Enhanced}
  transformer with rotary position embedding}.
\newblock \emph{Neurocomputing}, 568.

\bibitem[{Telgarsky(2015)}]{telgarsky2015representation}
Matus Telgarsky. 2015.
\newblock \href {https://arxiv.org/abs/1509.08101} {Representation benefits of
  deep feedforward networks}.
\newblock \emph{arXiv preprint arXiv:1509.08101}.

\bibitem[{Vaswani et~al.(2017)Vaswani, Shazeer, Parmar, Uszkoreit, Jones,
  Gomez, Kaiser, and Polosukhin}]{vaswani2017attention}
Ashish Vaswani, Noam Shazeer, Niki Parmar, Jakob Uszkoreit, Llion Jones,
  Aidan~N. Gomez, Lukasz Kaiser, and Illia Polosukhin. 2017.
\newblock \href
  {https://proceedings.neurips.cc/paper/2017/hash/3f5ee243547dee91fbd053c1c4a845aa-Abstract.html}
  {Attention is all you need}.
\newblock In \emph{Advances in Neural Information Processing Systems}.

\bibitem[{Virtanen et~al.(2020)Virtanen, Gommers, Oliphant, Haberland, Reddy,
  Cournapeau, Burovski, Peterson, Weckesser, Bright, {van der Walt}, Brett,
  Wilson, Millman, Mayorov, Nelson, Jones, Kern, Larson, Carey, Polat, Feng,
  Moore, {VanderPlas}, Laxalde, Perktold, Cimrman, Henriksen, Quintero, Harris,
  Archibald, Ribeiro, Pedregosa, {van Mulbregt}, and {SciPy 1.0
  Contributors}}]{SciPy}
Pauli Virtanen, Ralf Gommers, Travis~E. Oliphant, Matt Haberland, Tyler Reddy,
  David Cournapeau, Evgeni Burovski, Pearu Peterson, Warren Weckesser, Jonathan
  Bright, St{\'e}fan~J. {van der Walt}, Matthew Brett, Joshua Wilson, K.~Jarrod
  Millman, Nikolay Mayorov, Andrew R.~J. Nelson, Eric Jones, Robert Kern, Eric
  Larson, C~J Carey, {\.I}lhan Polat, Yu~Feng, Eric~W. Moore, Jake
  {VanderPlas}, Denis Laxalde, Josef Perktold, Robert Cimrman, Ian Henriksen,
  E.~A. Quintero, Charles~R. Harris, Anne~M. Archibald, Ant{\^o}nio~H. Ribeiro,
  Fabian Pedregosa, Paul {van Mulbregt}, and {SciPy 1.0 Contributors}. 2020.
\newblock \href {https://doi.org/10.1038/s41592-019-0686-2} {{SciPy} 1.0:
  Fundamental algorithms for scientific computing in {Python}}.
\newblock \emph{Nature Methods}, 17:261--272.

\bibitem[{von Oswald et~al.(2023{\natexlab{a}})von Oswald, Niklasson, Randazzo,
  Sacramento, Mordvintsev, Zhmoginov, and Vladymyrov}]{oswald23transformers}
Johannes von Oswald, Eyvind Niklasson, Ettore Randazzo, Jo\~ao Sacramento,
  Alexander Mordvintsev, Andrey Zhmoginov, and Max Vladymyrov.
  2023{\natexlab{a}}.
\newblock \href {https://proceedings.mlr.press/v202/von-oswald23a.html}
  {Transformers learn in-context by gradient descent}.
\newblock In \emph{International Conference on Machine Learning}.

\bibitem[{von Oswald et~al.(2023{\natexlab{b}})von Oswald, Niklasson, Schlegel,
  Kobayashi, Zucchet, Scherrer, Miller, Sandler, Vladymyrov, Pascanu, and
  Sacramento}]{von2023uncovering}
Johannes von Oswald, Eyvind Niklasson, Maximilian Schlegel, Seijin Kobayashi,
  Nicolas Zucchet, Nino Scherrer, Nolan Miller, Mark Sandler, Max Vladymyrov,
  Razvan Pascanu, and Jo\~ao Sacramento. 2023{\natexlab{b}}.
\newblock \href {https://arxiv.org/abs/2309.05858} {Uncovering
  mesa-optimization algorithms in transformers}.
\newblock \emph{arXiv preprint arXiv:2309.05858}.

\bibitem[{Wang and Xue(2023)}]{wang2024state}
Shida Wang and Beichen Xue. 2023.
\newblock \href
  {https://proceedings.neurips.cc/paper_files/paper/2023/hash/ea8608c6258450e75b3443ec8022fb2e-Abstract-Conference.html}
  {State-space models with layer-wise nonlinearity are universal approximators
  with exponential decaying memory}.
\newblock In \emph{Advances in Neural Information Processing Systems}.

\bibitem[{Wang et~al.(2023)Wang, Chauhan, Wang, and
  Hsieh}]{wang2024universality}
Yihan Wang, Jatin Chauhan, Wei Wang, and Cho-Jui Hsieh. 2023.
\newblock \href
  {https://proceedings.neurips.cc/paper_files/paper/2023/hash/eef6aecfe050b556c6a48d9c16b15558-Abstract-Conference.html}
  {Universality and limitations of prompt tuning}.
\newblock \emph{Advances in Neural Information Processing Systems}.

\bibitem[{Wei et~al.(2021)Wei, Bosma, Zhao, Guu, Yu, Lester, Du, Dai, and
  Le}]{wei2021finetuned}
Jason Wei, Maarten Bosma, Vincent Zhao, Kelvin Guu, Adams~Wei Yu, Brian Lester,
  Nan Du, Andrew~M Dai, and Quoc~V Le. 2021.
\newblock \href {https://openreview.net/forum?id=gEZrGCozdqR} {Finetuned
  language models are zero-shot learners}.
\newblock In \emph{International Conference on Learning Representations}.

\bibitem[{Weiss et~al.(2021)Weiss, Goldberg, and Yahav}]{weiss2021thinking}
Gail Weiss, Yoav Goldberg, and Eran Yahav. 2021.
\newblock \href {http://proceedings.mlr.press/v139/weiss21a.html} {Thinking
  like transformers}.
\newblock In \emph{International Conference on Machine Learning}.

\bibitem[{Xie et~al.(2022)Xie, Raghunathan, Liang, and Ma}]{xie2021explanation}
Sang~Michael Xie, Aditi Raghunathan, Percy Liang, and Tengyu Ma. 2022.
\newblock \href {https://openreview.net/forum?id=RdJVFCHjUMI} {An explanation
  of in-context learning as implicit {Bayesian} inference}.
\newblock In \emph{International Conference on Learning Representations}.

\bibitem[{Yun et~al.(2019)Yun, Bhojanapalli, Rawat, Reddi, and
  Kumar}]{yun2019transformers}
Chulhee Yun, Srinadh Bhojanapalli, Ankit~Singh Rawat, Sashank Reddi, and Sanjiv
  Kumar. 2019.
\newblock \href {https://arxiv.org/abs/1912.10077} {Are transformers universal
  approximators of sequence-to-sequence functions?}
\newblock In \emph{International Conference on Learning Representations}.

\bibitem[{Zhang and Sennrich(2019)}]{zhang2019root}
Biao Zhang and Rico Sennrich. 2019.
\newblock \href
  {https://proceedings.neurips.cc/paper/2019/hash/1e8a19426224ca89e83cef47f1e7f53b-Abstract.html}
  {Root mean square layer normalization}.
\newblock In \emph{Advances in Neural Information Processing Systems}.

\bibitem[{Zhang et~al.(2023)Zhang, Frei, and Bartlett}]{zhang2023trained}
Ruiqi Zhang, Spencer Frei, and Peter~L Bartlett. 2023.
\newblock \href {https://arxiv.org/abs/2306.09927} {Trained transformers learn
  linear models in-context}.
\newblock \emph{arXiv preprint arXiv:2306.09927}.

\bibitem[{Zhou et~al.(2024)Zhou, Bradley, Littwin, Razin, Saremi, Susskind,
  Bengio, and Nakkiran}]{zhou2023algorithms}
Hattie Zhou, Arwen Bradley, Etai Littwin, Noam Razin, Omid Saremi, Josh
  Susskind, Samy Bengio, and Preetum Nakkiran. 2024.
\newblock \href {https://arxiv.org/abs/2310.16028} {What algorithms can
  transformers learn? {A} study in length generalization}.
\newblock In \emph{International Conference on Learning Representations}.

\bibitem[{Ziyin et~al.(2020)Ziyin, Hartwig, and Ueda}]{ziyin2020neural}
Liu Ziyin, Tilman Hartwig, and Masahito Ueda. 2020.
\newblock \href
  {https://proceedings.neurips.cc/paper/2020/hash/1160453108d3e537255e9f7b931f4e90-Abstract.html}
  {Neural networks fail to learn periodic functions and how to fix it}.
\newblock In \emph{Advances in Neural Information Processing Systems}.

\end{thebibliography}
\bibliographystyle{acl_natbib}

\appendix

\newpage

\section{Computation Graph Debranching Rules}
\label{sec:debranching_rules}
We convert the computation DAG resulting from the LSRL program into a path program by attending to the first node whose output is the input for multiple other nodes, i.e., the first branching node.

\paragraph{Preparation step.}
Before we even start debranching we first pre-process the graph by fusing consecutive nodes of the same type together.
The specific rules are:
\begin{itemize}
    \item If a \Lin node is followed by a single other \Lin node, then fuse them together. This follows directly from the classical result that composing linear functions is a linear function.
    \item If a \ReLU node is followed by another \ReLU node, we can drop one of them as \ReLU is idempotent.
    \item If a \Lin is followed by a \LinState, we can subsume the weight matrix $\bm A$ of the linear node in the $\bm B$ matrix of the \LinState, and the bias $\bm b$ of the \Lin node in the bias $\bm b$ of the \LinState.
    \item If all inputs of a \Concat node are the same, then this node only duplicates the input and hence can be safely replaced with a \Lin layer.
\end{itemize}

The debranching process goes through the following cases in order.
And iterates until there are no branching nodes left, in other words, until the graph has become a path graph.
We will refer to the nodes whose input is the branching node as \emph{subsequent nodes}.

\paragraph{Case 1A: If all subsequent nodes are \Multi.}
As all \Multi nodes that have the same input (the branching node) they must all be  producing the exact same output.
Hence, only one can be kept.
This removes one branch.

\paragraph{Case 1B: If subsequent nodes are a combination of \Multi and other nodes.}
We add a single \Lin layer that acts as a bypass for the non-\Multi nodes using the fact that multiplicatin by 1 is identity.
This is followed by a single \Multi layer.
We then add \Slice operators between the new \Lin layer and the non-\Multi nodes.
This keeps the number of branches unchanged but removes the \Multi node and the new branch can be handled by the other rules.

\paragraph{Case 2: All subsequent nodes are \LinState.}
\LinState nodes can be fused into a single \LinState node by combining their states and update matrices.
As each \LinState may have different subsequent nodes itself, we add \Slice nodes to extract the respective subspaces of the state.
This keeps the number of branches unchanged but puts the graph into Case 5A.

\paragraph{Case 3: All subsequent nodes are \ReLU.}
We can replace them by a single \ReLU node.
This removes one branch.

\paragraph{Case 4: All subsequent nodes are \Concat.}
One complication is that \Concat nodes can depend on other \Concat nodes.
So, we will restrict ourselves at this step by only treating the \Concat nodes that depend only on the branch node directly by replacing them with a single \Lin node.
The rest will be handled by the \Lin and \Concat case (Case 10) or the only \Lin case (Cases 5A and 5B).
See the following example:
\begin{center}
    \begin{minipage}[c]{0.32\textwidth}
        \begin{tikzpicture}[node distance=0.5cm and 0.5cm,
            every node/.style={draw, rectangle},
            line/.style={draw, thick, -Stealth}]
            
            \node (branch) {\texttt{BranchNode}};
            \node (concat1) [below right=of branch] {\Concat};
            \node (concat2) [below left=of concat1] {\Concat};
            
            \draw[line] (branch) to[bend right=25] (concat1);
            \draw[line] (branch) to[bend left=25] (concat1);
            \draw[line] (branch) -- (concat2);
            \draw[line] (concat1) -- (concat2);
        \end{tikzpicture}
    \end{minipage}
$\implies$
    \begin{minipage}[c]{0.32\textwidth}
        \begin{tikzpicture}[node distance=0.5cm and 0.5cm,
            every node/.style={draw, rectangle},
            line/.style={draw, thick, -Stealth}]
            
            \node (branch) {\texttt{BranchNode}};
            \node (concat1) [below right=of branch] {\Lin};
            \node (concat2) [below left=of concat1] {\Concat};
            
            \draw[line] (branch) to (concat1);
            \draw[line] (branch) -- (concat2);
            \draw[line] (concat1) -- (concat2);
        \end{tikzpicture}
    \end{minipage}
\end{center}
Hence, this operation either reduces the number of branches by one or will be followed by a case that reduces the number of branches.

\paragraph{Case 5A: Only \Lin nodes and they are all \Slice{}s.}
This is one of the more challenging cases.
While the \Slice nodes are simply \Lin nodes with special structure, we cannot treat them like standard \Lin nodes (see Case 5B).
While we can merge them into a single \Lin node, we will then need further \Slice{}s to extract the relevant subspaces for the subsequent nodes.
Therefore, we would be simply replacing \Slice nodes with \Slice nodes.
Instead, we use the observation that \Slice nodes can be fused with subsequent \Lin and \LinState nodes and can be pushed after \ReLU and \Concat nodes.
Therefore we treat each subsequent node differently, depending on its type:
\begin{itemize}
    \item If there are \Multi nodes after any of the \Slice nodes, they can all be fused into a single \Lin node followed by a single \Multi node.
    \item If there are \Lin or \LinState nodes after any of the \Slice nodes, the \Slice{}s can be fused with the $\bm A$ matrix of the \Lin nodes and the $\bm B$ matrix of the \LinState nodes.
        This uses the fact that composing linear functions results in a linear function.
    \item If there is a \ReLU after a \Slice node, their position can be switched without changing the nodes. That is because \ReLU commutes with linear operations with $\bm b=\bm 0$ and $\bm A$ with non-negative eigenvalues as is the case for \Slice nodes.
    \item If there is a \Concat node after a \Slice node, we can similarly push the \Slice as a new \Lin node after the \Concat.
\end{itemize}

This step does not reduce the number of branching nodes but prepares the graph for a removal, with the specific case depending on the remaining nodes.

\paragraph{Case 5B: Only \Lin nodes and they are not all \Slice{}s.}
We can combine them into a single \Lin node and then add \Slice{}s to extract the relevant subspaces for the subsequent nodes.
These \Slice{}s can then be pushed into the next operations using Case 5A.

\paragraph{Case 6: Both \LinState nodes and other nodes.}
If both \LinState nodes and other nodes are present, we can pass through the other variables with dummy \LinState variables using zero matrices for $\bm A$ and identities for $\bm B$.
Then, Case 2 can be used to fuse all the \LinState variables together.

\paragraph{Case 7A: Only \Lin and \ReLU nodes where all \Lin nodes are followed by only one node which is a \ReLU.}
If we add \Lin bypasses to the \ReLU{}s we will have only \Lin nodes left.
Each one of them would be followed by a \ReLU. 
Hence, Case 5B can be first applied, followed by Case 3.

\paragraph{Case 7B: Only \Lin and \ReLU nodes where some \Lin nodes are not followed by only one node which is a \ReLU.}
In this case we cannot apply the above strategy.
Instead, we fuse the \ReLU{}s by placing \ReLU-based bypasses before the \Lin nodes.
We do this in a similar spirit to \Cref{eq:linear_to_nonlinear}, by splitting the positive and negative components and treating them separately.
See \Cref{sec:relu_identity} for the LSRL implementation.
Our DAG will then be in Case 7A first, then Case 5B, and, finally, in Case 3.

\paragraph{Case 8: Only \Lin and \Concat nodes.}
We add \Lin bypasses for the \Concat nodes which can then be merged using Case 5B and then Case 5A.

\paragraph{Case 9: Only \ReLU and \Concat nodes.}
Same strategy as for Case 8 but with \ReLU bypasses.

\paragraph{Case 10: Only \Lin, \ReLU or \Concat nodes.}
We introduce \ReLU bypasses to all \Concat nodes and to the \Lin branches which are not immediately followed by a \ReLU.
This will be followed by applying Case 5B and then Case 3.

The above 13 cases cover all possible branching configurations.
After repeated application, they reduce any DAG corresponding to an LSRL program to a path graph that can be compiled to one of the recurrent models in \Cref{sec:preliminaries}.

\section{Error Bound on the Approximation Scheme for Continuous Functions}
\label{sec:error_bound}

In \Cref{sec:approximating_cont_functions} we outlined a strategy to perform universal in-context approximation for continuous functions with Linear RNNs.
The full program is in \Cref{lst:continous_approximation} and an illustration of the scheme is presented in \Cref{fig:vec2vec}.
In \Cref{sec:approximating_cont_functions} we claimed that the prompt length required to approximate an $L$-Lipschitz function $f$ (w.r.t. the $\ell_2$ norm) to precision $\epsilon$ is $N=(\nicefrac{2\epsilon}{L\sqrt{d_\text{in}}})^{\texttt{-}d_\text{in}} = \mathcal O ( \epsilon^{\texttt{-}d_\text{in}})$.
The present appendix offers the formal proof of this claim.

The program in \Cref{lst:continous_approximation} approximates the value of a function $\bm y=f(\bm q)$ with the value $\bar{\bm y}$ at the centre $\bm c$ of the cell that contains $\bm q$. Therefore, the error of our approximation is the maximum difference between $f(\bm q)$ and $f(\bm c)$: $\|f(\bm q)-f(\bm c)\|_2$.
First, as the length of each side of the cell is $\delta$, that means that $\|\bm q - \bm c\|_\infty \leq \nicefrac{\delta}{2}$.
Thus, $\|\bm q - \bm c\|_2 \leq \nicefrac{\sqrt{d_\text{in}}\delta}{2}$.
Therefore, thanks to $f$ being $L$-Lipschitz we get:
\begin{align*}
    \|f(\bm q)-f(\bm c)\|_2
    &\leq \frac{\delta L \sqrt{d_\text{in}}}{2}.
\end{align*}
If we want to upper bound this approximation error by $\epsilon$, we need to have $\delta$ small enough:
\begin{equation*}
    \delta \leq \frac{2 \epsilon}{L \sqrt{d_\text{in}}}.
\end{equation*}
Finally, as the number of cells we need to cover the whole domain is $N=(\nicefrac{1}{\delta})^{d_\text{in}}$, this corresponds to us needing sufficiently long prompt:
\begin{equation*}
    N
    \geq 
    \left(\frac{1}{\delta}\right)^{d_\text{in}} 
    \geq
    \left(\frac{L \sqrt{d_\text{in}}}{2 \epsilon}\right)^{d_\text{in}}.
\end{equation*}
Therefore, if we want our approximation to have error at most $\epsilon$ anywhere in the domain, we need a prompt of length at least $(\nicefrac{L \sqrt{d_\text{in}}}{2 \epsilon})^{d_\text{in}}$.

\section{Gated RNNs are GRU models} 
\label{sec:GRU}
A GRU layer \citep{cho2014learning} with input $\bm a_t\in\RR^{d_\texttt{in}}$ and hidden state $\bm h_{t-1}\in\RR^{d_\texttt{hidden}}$, and output $\bm h_t\in\RR^{d_\texttt{hidden}}$ can be described as follows:
\begin{align}
    \bm z_t &= \texttt{Sigmoid}(\bm W_z \bm a_t + \bm U_z \bm h_{t-1} + \bm b_z), && \text{(update gate vector)}\\
    \bm r_t &= \texttt{Sigmoid}(\bm W_r \bm a_t + \bm U_r \bm h_{t-1} + \bm b_r), &&\text{(reset gate vector)} \\
    \hat{\bm h}_t &= \texttt{tanh}(\bm W_h \bm a_t + \bm U_h(\bm r_t  \odot \bm h_{t-1}) + \bm b_h), && \text{(candidate activation vector)} \label{eq:gru_cav} \\
    \bm h_t &= (1 - \bm z_t) \odot \bm h_{t-1} + \bm z_t \odot \hat{\bm h}_t, &&\text{(output vector)} \label{eq:gru_output}
\end{align}

In this section, we show a conversion of  a single Gated RNN layer (\Cref{eq:gated_rnn}) to $k+2$ GRU layers.
Here, $k$ is the number of layers in the $\gamma$ and $h$ MLPs in \Cref{eq:gated_rnn}. 
We first show that a single GRU layer can be used to compute the updated state $\bm s_{t}$ and the output of the first layer of $\gamma$ when applied to $\bm x_t$. 
Then, every pair of single layers of $\gamma(\bm x_t)$ and $\phi(\bm s_t)$ can be represented as an individual GRU layer. 
Finally, a single layer can be used to compute the element-wise multiplication $\gamma(\bm x_t)\odot \phi(\bm s_t)$.
For simplicity, we assume the \texttt{Sigmoid} and \texttt{tanh} nonlinearities are replaced by \ReLU{}s.
If not, they can each be approximated with MLPs and hence also with additional GRU layers. 
Additionally, for convenience we will assume $d_\texttt{in} = d_\texttt{hidden}$. 

\subsection{Representing the state update as a GRU layer}\label{sec:gru_recurrent_layer}
For this layer we set $\bm b_z = \bm 1$, $\bm W_z = \bm 0$ , $\bm U_z = \bm 0$ giving $\bm z_t = \bm 1$. 
Similarly, we set $\bm b_r = \bm 1$, $\bm W_r = \bm 0$ , $\bm U_r = \bm 0$ giving $\bm r_t = \bm 1$. 
Thus, \Cref{eq:gru_cav} reduces to:
\begin{equation}
    \bm{\hat{h}}_t = \sigma(\bm W_h \bm a_t + \bm U_h\bm h_{t-1} + \bm b_h), \label{eq:gru_mixing_layer}
\end{equation}
Setting 
$\bm a_t = \begin{bmatrix} \bm 0 \\ \bm x_t  \end{bmatrix}$, where $\bm x_t\in\RR^{d_\texttt{in}/2}$, 
$\bm h_{t-1} = \begin{bmatrix} \bm s_{t-1} \\ \bm 0 \end{bmatrix}$, where $\bm s_{t-1}\in\RR^{d_\texttt{hidden}/2}$, 
$\bm W_h = \begin{bmatrix} \bm 0 & \bm B \\ \bm 0 & \bm I\end{bmatrix}$, 
$\bm U_h = \begin{bmatrix} \bm A & \bm 0 \\  \bm 0 & \bm 0 \end{bmatrix}$, 
$\bm b_h =  \begin{bmatrix} \bm b \\  -\bm k_{lb} \end{bmatrix}$, where $\bm k_{lb}$ is a vector where every element in $\bm k$ is a lower bound on $\bm x_t$. 
results in \Cref{eq:gru_output} becoming:
\begin{align}
    \bm h_t &= \sigma\left(\begin{bmatrix} \bm 0 & \bm B \\ \bm 0 & \bm I\end{bmatrix}\begin{bmatrix} \bm 0 \\ \bm x_t  \end{bmatrix} + \begin{bmatrix} \bm A & \bm 0 \\  \bm 0 & \bm 0 \end{bmatrix}\begin{bmatrix} \bm s_{t-1} \\ \bm 0 \end{bmatrix} + \begin{bmatrix} \bm b \\  \bm -\bm k_{lb}  \end{bmatrix}\right)
    = \begin{bmatrix} \sigma(\bm A \bm s_{t-1} + \bm B \bm x_t + \bm b) \\ \sigma(\bm x_t -\bm k_{lb}) \end{bmatrix} = \begin{bmatrix} \sigma(\bm s_t) \\ \bm x_t -\bm k_{lb} \end{bmatrix}.
    \label{eq:gru_mixing_layer_2}
\end{align}
\emph{Note: if we do not want to assume a compact domain for $\bm x_t$, it would be possible to use the same trick as in Equation (\ref{eq:linear_to_nonlinear}) rather than subtracting $\bm k$ in this layer and adding in the next. However, we omit this approach for clarity of presentation.}

\subsection{Representing each MLP layer as a GRU layer}\label{sec:gru_mlp_layer}
In these layers, similarly to the recurrent layer, we set $\bm b_z = \bm 1$, $\bm W_z = \bm 0$ , $\bm U_z = \bm 0$ giving $\bm z_t = \bm 1$. 
In the same way, we set $\bm b_r = \bm 1$, $\bm W_r =  \bm 0$ , $\bm U_r = \bm 0$ giving $\bm r_t = \bm 1$. 
Here, however, we set
$\bm W_h = \begin{bmatrix} \bm W_{h_i} & \bm 0 \\ \bm 0 & \bm W_{\gamma_i} \end{bmatrix}$, 
$\bm U_h = \bm 0$ and 
$\bm b_h =  \begin{bmatrix} \bm b_{h_i} \\  \bm b_{\gamma_i} \end{bmatrix}$, except for the first of such layer where $\bm b_h =  \begin{bmatrix} \bm b_{h_i} \\ \bm b_{\gamma_i} + \bm W_{\gamma_i} \bm k_{lb} \end{bmatrix}$. 
Thus, for an input $\bm a_t = \begin{bmatrix} \bm a_{1,t} \\ \bm a_{2,t}  \end{bmatrix}$ the layer output (\Cref{eq:gru_output}) for layer $i$ is:
\begin{equation}
    \bm h_t = \sigma\left(\begin{bmatrix} \bm W_{\phi_i} & \bm 0 \\ \bm 0 & \bm W_{\gamma_i} \end{bmatrix}\begin{bmatrix} \bm a_{1,t} \\ \bm a_{2,t}  \end{bmatrix}  + \begin{bmatrix} \bm b_{\phi_i} \\  \bm b_{\gamma_i} \end{bmatrix}\right)
    = \begin{bmatrix} \phi_i(\bm a_{1,t}) \\ \gamma_i(\bm a_{1,t})\end{bmatrix}.
    \label{eq:gru_mlp_layer}
\end{equation}
Here, $\phi_i$ and $\gamma_i$ are the $i$-th layers (including the \ReLU) of respectively $\phi$ and $\gamma$ in \Cref{eq:gated_rnn}. 

\subsection{Representing the multiplicative gating with a single GRU layer}\label{sec:gru_mixing_layer}
The only thing left is to model the element-wise multiplication of the outputs of $\phi$ and $\gamma$ in \Cref{eq:gated_rnn}.
We do this using a GRU layer with $\bm b_z = \bm 0$, $\bm W_z = \bm 0$, $\bm U_z =  \begin{bmatrix} \bm 0 & \bm 0 \\ \bm 0 & \bm I \end{bmatrix}$. We set $\bm b_r = \bm 0$, $\bm W_r = \bm 0$ , $\bm U_r = \bm 0$ giving $\bm r_t  = \bm 0$.
We also set $\bm b_h = \bm 0$, $\bm W_h = \begin{bmatrix} \bm 0 & \bm 0 \\ \bm I & \bm 0 \end{bmatrix}$, $\bm U_h = \bm 0$. 
Thus, for an input $\bm a_t = \begin{bmatrix} \bm a_{1,t} \\ \bm a_{2,t}  \end{bmatrix}$, the output $\bm h_t$ (\Cref{eq:gru_output}) of this GRU layer becomes:
\begin{equation}
    \bm h_t = \sigma\left(\begin{bmatrix} \bm 0 & \bm 0 \\ \bm I & \bm 0 \end{bmatrix}\begin{bmatrix} \bm a_{1,t} \\ \bm a_{2,t}  \end{bmatrix}\right) \odot \begin{bmatrix} \bm 0 & \bm 0 \\ \bm 0 & \bm I \end{bmatrix}\begin{bmatrix} \bm a_{1,t} \\ \bm a_{2,t}  \end{bmatrix}
         = \begin{bmatrix} \bm 0 \\ \sigma(\bm a_{1,t}) \odot \bm a_{2,t}\end{bmatrix}.
    \label{eq:gru_multi_layer}
\end{equation}
If $\bm a_t$ is the output of a GRU layer constructed as in \Cref{eq:gru_mlp_layer} (as is in our case), then it must be non-negative.
This is due to the \ReLU application in \Cref{eq:gru_mlp_layer}.
Hence, the application of another \ReLU to $\bm a_{1,t}$ in \Cref{eq:gru_multi_layer} can be safely removed as \ReLU is idempotent and \Cref{eq:gru_multi_layer} simplifies to
\begin{equation}
    \bm h_t = \begin{bmatrix} \bm 0  \\ \bm a_{1,t} \odot \bm a_{2,t} \end{bmatrix}.
    \label{eq:gru_multi_layer_simplified}
\end{equation}
Thus, this construction computes element-wise multiplication of $\bm a_{1,t}$ and $\bm a_{2,t}$.

\subsection{Composing the operations to model a single Gated RNN layer}
In order to represent \Cref{eq:gated_rnn}, we use one GRU layer for the recurrence (as described in \Cref{sec:gru_recurrent_layer}), followed by $k$ GRU layers modelling a pair of  the $k$ MLP layers of $\phi$ and $\gamma$ (\Cref{sec:gru_mlp_layer}), completed with  a single mixing layer (\Cref{sec:gru_mixing_layer}). 
This stack of $k+2$ layers models exactly the Gated  RNN layer (\Cref{eq:gated_rnn}):
\begin{align*}
        \bm s_t &= \sigma\left(\bm A \begin{bmatrix} \bm 0 \\ \bm s_{t-1} \end{bmatrix} + \bm B\begin{bmatrix} \bm x_t  \\ \bm 0 \end{bmatrix} + \bm b\right) \\
        \bm y_t &= \begin{bmatrix} \bm 0 \\  \gamma(\bm x_t) \odot \phi(\bm s_t) \end{bmatrix},
\end{align*}

With this, we have shown that any Gated RNN (\Cref{eq:gated_rnn}) can be expressed as a GRU-based model.
Hence, the two universal approximation programs in \Cref{lst:continous_approximation,lst:tok2tok_approximation} can be implemented also in GRU-based models.
Thus, the GRU architecture can also be a universal in-context approximator. 

\section{Gated RNNs are LSTMs}
\label{sec:lstms}

A single LSTM layer \citep{hochreiter1997long, gers2000learning} with input $\bm a_t\in\RR^{d_\texttt{in}}$, hidden state $\bm h_{t-1}\in\RR^{d_\texttt{hidden}}$, candidate memory cell $\tilde{\bm c_t} \in\RR^{d_\texttt{hidden}}$, memory cell $\bm c_t\in\RR^{d_\texttt{hidden}}$ and layer output $\bm h_t\in\RR^{d_\texttt{hidden}}$ can be expressed as:
\begin{align}\bm f_t &= \texttt{Sigmoid}(\bm W_f \bm a_t + \bm U_f \bm h_{t-1} + \bm b_f), && \text{(forget gate vector)} \\ \bm i_t &= \texttt{Sigmoid}(\bm W_i \bm a_t + \bm U_i \bm h_{t-1} + \bm b_i), && \text{(input gate vector)} \\ \bm o_t &= \texttt{Sigmoid}(\bm W_o \bm a_t + \bm U_o \bm h_{t-1} + \bm b_o), && \text{(output gate vector)} \\  \tilde{ \bm c}_t &= \texttt{tanh}(\bm W_c \bm a_t + \bm U_c \bm h_{t-1} + \bm b_c), && \text{(candidate cell vector)}\\ \bm  c_t &= \bm f_t \odot \bm c_{t-1} + \bm i_t \odot \tilde{\bm c}_t, && \text{(memory cell vector)} \\ \bm h_t &= \bm o_t \odot \texttt{tanh}(\bm c_t), && \text{(output vector)}
\end{align}
where $\bm h_0 = \bm 0$ and $\bm c_0 = \bm 0$ .

In a way analogous to \Cref{sec:GRU}, we show that a single layer of a gated RNN (\Cref{eq:gated_rnn}) can be expressed using $k+2$ LSTM layers, where $k$ is the maximum depth of either of the MLP networks $\phi$ or $\gamma$.
We again follow the setup of replacing all \texttt{Sigmoid} and \texttt{tanh} activation functions with \ReLU activations which we denote $\sigma$ and we again assume that $d_\texttt{in} = d_\texttt{hidden}$.
The set up follows the same structure as in \Cref{sec:GRU}.
First, we show that the non-linear state update computing $\bm s_t$ can be expressed as a single LSTM layer. 
We then show that we can represent the layers in MLP networks $\gamma(\bm x_t)$ and $\phi(\bm s_t)$ using single LSTM layers. 
Finally, a single layer can compute the Hadamard product between $\gamma(\bm x_t)$ and $\phi(\bm s_t)$.
Therefore, any Gated RNN with \ReLU activations can be expressed as a LSTM with \ReLU activations. 

For clarity of the exposition, we once again assume that our inputs belong to a compact domain $\mathcal{X}$ of real vectors. 
This implies that the set is bounded and, in particular, that we can find a vector $\bm k_{lb}$ such that $\bm k_{lb,i} \leq ( \bm x_t)_i$ for $i \in [d_\texttt{in}]$ for all $\bm x_t \in \mathcal{X}$.
In other words, we have $\bm (\bm x_t - \bm k_{lb})_i \geq 0$ for for $i \in 1,\ldots,d_\texttt{in}$. 
We will make use of this fact several times when dealing with \ReLU activations.

\subsection{Representing the state update as an LSTM layer}\label{sec:lstm_recurrent_layer}
We first represent the non-linear state update in \cref{eq:gated_rnn} using a single layer of an LSTM. In particular, we set $\bm W_f = \bm 0$, $\bm U_f = \bm 0$ and $\bm b_f = \bm 0$ so that $\bm f_t = \bm 0$. We also set $\bm W_i = \bm 0$, $\bm U_i = \bm 0$, $\bm b_i = \bm 1$ and $\bm W_c = \bm 0$, $\bm U_c = \bm 0$, $\bm b_c = \bm 1$. 
This results in $\bm i_t = \bm 1$ and $\bm \tilde{\bm c}_t = \bm 1$. We see from this that the LSTM layer with these weight settings reduces to 
\begin{equation}
\bm h_t = \bm o_t = \sigma(\bm W_o \bm a_t + \bm U_o \bm h_{t-1} + \bm b_o).
\end{equation}
We now set $\bm a_t = \begin{bmatrix} \bm 0 \\ \bm x_t  \end{bmatrix}$, where $\bm x_t\in\RR^{d_\texttt{in}/2}$, 
$\bm h_{t-1} = \begin{bmatrix} \bm s_{t-1} \\ \bm 0 \end{bmatrix}$, where $\bm s_{t-1}\in\RR^{d_\texttt{hidden}/2}$, 
$\bm W_o = \begin{bmatrix} \bm 0 & \bm B \\ \bm 0 & \bm I\end{bmatrix}$, 
$\bm U_o = \begin{bmatrix} \bm A & \bm 0 \\  \bm 0 & \bm 0 \end{bmatrix}$, 
$\bm b_o =  \begin{bmatrix} \bm b \\  -\bm k_{lb} \end{bmatrix}$ so that 
\begin{align}
    \bm h_t &= \sigma\left(\begin{bmatrix} \bm 0 & \bm B \\ \bm 0 & \bm I\end{bmatrix}\begin{bmatrix} \bm 0 \\ \bm x_t  \end{bmatrix} + \begin{bmatrix} \bm A & \bm 0 \\  \bm 0 & \bm 0 \end{bmatrix}\begin{bmatrix} \bm s_{t-1} \\ \bm 0 \end{bmatrix} + \begin{bmatrix} \bm b \\  -\bm k_{lb} \end{bmatrix}\right)
         = \begin{bmatrix} \sigma(\bm A \bm s_{t-1} + \bm B \bm x_t + \bm b) \\ \sigma(\bm x_t - \bm k_{lb}) \end{bmatrix} = \begin{bmatrix}
             \bm s_t \\ \bm x_t -\bm k_{lb}
         \end{bmatrix}.
    \label{eq:lstm_mixing_layer_2}
\end{align}

\subsection{Representing each MLP layer as an LSTM layer}\label{sec:lstm_mlp_layer}
Now we want to use an LSTM layers to model the MLP layers of both $\gamma$ and $\phi$ simultaneously. We set $\bm W_f = \bm 0$, $\bm U_f = \bm 0$, $\bm b_f = \bm 0$ and $\bm W_i = \bm 0$, $\bm U_i = \bm 0$, $\bm b_i = \bm 1$ and $\bm W_c = \bm 0$, $\bm U_c = \bm 0$, $\bm b_c = \bm 1$ as before. We make a change for these LSTM layers by setting
$\bm W_o = \begin{bmatrix} \bm W_{\phi_i} & \bm 0 \\ \bm 0 & \bm W_{\gamma_i} \end{bmatrix}$, 
$\bm U_o = \bm 0$ and 
$\bm b_o =  \begin{bmatrix} \bm b_{\phi_i} \\  \bm b_{\gamma_i} \end{bmatrix}$, except for the first layer where $\bm b_\phi =  \begin{bmatrix} \bm b_{\phi_1} \\  \bm b_{\gamma_1} + \bm W_{\gamma_1}\bm k \end{bmatrix}$.
Thus, for an input $\bm a_t = \begin{bmatrix} \bm a_{1,t} \\ \bm a_{2,t}  \end{bmatrix}$ the layer output is:
\begin{equation}
    \bm h_t = \sigma\left(\begin{bmatrix} \bm W_{\phi_i} & \bm 0 \\ \bm 0 & \bm W_{\gamma_i} \end{bmatrix}\begin{bmatrix} \bm a_{1,t} \\ \bm a_{2,t}  \end{bmatrix}  + \begin{bmatrix} \bm b_{\phi_i} \\  \bm b_{\gamma_i} \end{bmatrix}\right)
    = \begin{bmatrix} \phi_i(\bm a_{1,t}) \\ \gamma_i(\bm a_{2,t})\end{bmatrix}.
\end{equation}
Here, $\phi_i$ and $\gamma_i$ again refer to the $i$-th layers (including the \ReLU) of respectively $\phi$ and $\gamma$ in \Cref{eq:gated_rnn}. 

Note that, without a loss of generality, if we have that $\phi$ has $m$ layers whereas $\gamma$ has $k$ with $m < k$, then we can also model this by simply adding additional layers to model additional layers for $\gamma$ whilst simply passing on $\phi$ unchanged. Specifically, we set set the weights to ensure that $\bm f_t =0 $ and that $\bm i_t$ and $\tilde{\bm c_t} $ are $\bm 1$ so that $\bm h_t = \bm o_t$. The input to this layer for $i > k$ is then given as $\bm a_t = \begin{bmatrix} \phi (\bm s_t) \\ \bm a_{2,t}  \end{bmatrix}$.  The we set the weights to compute $\bm o_t$ as
\begin{equation}
    \bm o_t = \sigma\left(\begin{bmatrix} \bm I & \bm 0 \\ \bm 0 &  \bm W_{\gamma_i} \end{bmatrix}\begin{bmatrix} \phi (\bm s_t) \\ \bm a_{2,t}  \end{bmatrix}  + \begin{bmatrix} \bm 0 \\  \bm b_{\phi_i} \end{bmatrix}\right)
    = \begin{bmatrix} \phi (\bm s_t)\\ \gamma_i(\bm a_{2,t})\end{bmatrix}.
\end{equation}

\subsection{Representing the multiplicative gating with an LSTM layer}
\label{sec:lstm_mixing_layer}
Finally, we model the element-wise multiplication of the outputs of $\phi$ and $\gamma$ in \Cref{eq:gated_rnn}. To do this we set the weights of the input gate and candidate cell vectors for the final layers of of $\gamma$ and $\phi$ to be as follows:
\begin{equation}
    \bm i_t = \sigma\left(\begin{bmatrix} \bm 0 & \bm 0 \\ \bm I & \bm 0 \end{bmatrix}\begin{bmatrix} \bm a_{1,t} \\ \bm a_{2,t}  \end{bmatrix}  + \begin{bmatrix} \bm 0 \\  \bm 0 \end{bmatrix}\right)
    = \begin{bmatrix} \bm 0 \\ \bm a_{1,t}  \end{bmatrix} 
\end{equation}
and
\begin{equation}
    \tilde{\bm c} = \sigma\left(\begin{bmatrix} \bm 0 & \bm 0 \\ \bm 0 & \bm I \end{bmatrix}\begin{bmatrix} \bm a_{1,t} \\ \bm a_{2,t}  \end{bmatrix}  + \begin{bmatrix} \bm 0 \\  \bm 0 \end{bmatrix}\right)
    = \begin{bmatrix} \bm 0 \\ \bm a_{2,t} \end{bmatrix}.
\end{equation}

Then by setting $\bm W_f = \bm 0$, $\bm U_f = \bm 0$, $\bm b_f = \bm 0$ and $\bm W_o = \bm 0$, $\bm U_o = \bm 0$, $\bm b_o = \bm 1$ to force $\bm f_t = \bm 0$ and $\bm o_t = \bm 1$, we get 
\begin{equation}
\label{eqn: lstm_prod}
    \bm y_t = \sigma(\bm c_t) = \begin{bmatrix} \sigma (\bm 0  \odot \bm 0) \\ \sigma (\bm a_{1,t} \odot \bm a_{2,t})  \end{bmatrix} = \begin{bmatrix} \bm 0 \\ \sigma (\bm a_{1,t} \odot \bm a_{2,t}) \end{bmatrix}.
\end{equation}

\subsection{Composing the operations to model a single Gated RNN layer}
To model the gated RNN described in \cref{eq:gated_rnn}, we again follow the same lines as described in \cref{sec:GRU}. In particular, we use one LSTM layer for the recurrent state updated as described in \cref{sec:lstm_recurrent_layer}. We then stack $k$ LSTM layers as described in \cref{sec:lstm_mlp_layer} to model the $k$ MLP layers of $\phi$ and $\gamma$ in parallel. We then use one final layer to both give the final MLP layer of $\phi$ and $\gamma$ and to compute their Hadamard product as set out in \cref{sec:lstm_mixing_layer} in order to match the output of the gated RNN in \cref{eq:gated_rnn}. 
Now, since we are working with $\sigma = \text{ReLU}$, both $\gamma(\bm x_t)$ and $\phi(\bm s_t)$ are positive and therefore so is their product. Hence, applying $\sigma$ to the product components in \cref{eqn: lstm_prod} leaves the the components invariant. Therefore, we output is 
\begin{equation}
    \bm y_t = \begin{bmatrix} \bm 0\\ \gamma(\bm x_t) \odot \phi (\bm s_t) \end{bmatrix},
\end{equation}
as required.

Hence, we have shown that a single layer of a gated RNN as described by \cref{eq:gated_rnn} can be represented using $k+2$ LSTM layers where $k $ is the maximum depth of $\phi$ and $\gamma$.
Therefore, once again, the two universal approximation programs in \Cref{lst:continous_approximation,lst:tok2tok_approximation} can also be implemented for LSTMs.
Hence, LSTM models are also universal approximators in the sense described in \Cref{sec:ua_linear_RNNs}.

\section{Gated Linear RNNs are Hawk/Griffin Models}
\label{sec:hawk_griffin}
A single residual block of a Hawk/Griffin model \citep{de2024griffin} consists of two components, a recurrent block for temporal mixing which makes use of a one-dimensional temporal convolution, as well as real-gated linear recurrent unit (RG-LRU) and a gated MLP block. Specifically,  we consider an input $\bm a_t\in\RR^{d_\texttt{in}}$, inputs to the blocks of dimensions $d_\texttt{in}$ and outputs from each block of dimensions $d_\texttt{in}$. Within blocks, all vectors have dimensionality $d_\texttt{hidden} = E d_\texttt{in}$, where $E$ is denotes an expansion factor.
Below, we formally describe the form of the recurrent and gated MLP blocks which are the two main components making up the residual blocks used for Hawk and Griffin.

\textbf{Recurrent block}. The recurrent block consists of two branches. The first applies a one-dimensional temporal convolution followed by a RG-LRU. The second branch simply performs a linear transformation followed by a non-linearity, i.e. applies a single layer of an MLP. 

Consider the first branch of the recurrent block with an input $\bm a_t$. The one-dimensional temporal convolution can be written as:
\begin{align}
   \bm a'_t &= \bm W_a \bm a_t, \\
    \bm g_t &= \texttt{GeLU}(\bm W_g \bm a_t + \bm b_g),  \\
    \bm M_t &=
        \begin{bmatrix}
\bm{a}'_{t-(d_\text{conv}-1)}, \ldots, \bm{a}'_{t-2}, \bm{a}'_{t-1}, \bm{a}'_t
\end{bmatrix}, \\
    \bm z_t &= \sum_{i=0}^{d_\text{conv}\texttt{-}1} \bm W_M[i] \bm M_t[t-i]  ~~+\bm b_\text{conv} && \text{(convolution with window size $d_{\text{conv}}$)},   
    \end{align}    
where $\bm{b}_{\text{conv}}$ is a bias vector and $\bm W_M = \begin{bmatrix}
    \tilde{\bm  B},  \tilde{\bm  A} \tilde{ \bm  B}, \tilde{\bm  A}^2 \tilde{\bm  B},  \cdots,  \tilde{\bm  A}^t \bm \tilde{\bm  B}, \cdots 
\end{bmatrix}$ is the convolutional kernel for the one-dimensional temporal convolution.

The output of this convolution is then fed into a RG-LRU. We can write this down concretely using as an input $\bm z_t$ from the one-dimensional convolution and with recurrent state $\bm h_t \in \RR^{d_{\texttt{model}}}$:
\begin{align}
     \bm r_t &= \texttt{Sigmoid}(\bm W_r \bm z_t + \bm b_r),   \\
     \bm i_t &= \texttt{Sigmoid}(\bm W_i \bm z_t + \bm b_i),  \\
    a &= \texttt{Sigmoid}(\Lambda), && (\Lambda \text{ a learnable parameter}) \\
     \bm a_t &= a^{c \bm r_t}, && (c=8 \text{ fixed scalar constant})\\
    \bm h_t &= \bm a_t \odot \bm h_{t-1} + \sqrt{1-\bm a_t^2} \odot ( \bm i_t \odot  \bm z_t). 
\end{align}

Now consider the second branch of the recurrent block. This performs a linear transformation followed by a non-linear activation:
\begin{align}
\bm g_t = \texttt{GeLU}(\bm W_g \bm a_t + \bm b_g). 
\end{align}

To get the final output of the recurrent block, we multiply the components of the vectors computed from each branch within the recurrent block and then perform a non-linear transformation: 
\begin{align}
    \bm h'_t &= \bm g_t \odot \bm h_t, \\
    \bm o_t &= \bm W_o \bm h'_t + \bm b_o . 
\end{align}

\textbf{Gated MLP block}. After passing through the recurrent block, we pass the output $\bm o _t$ into a gated MLP block. Again we have two branches, the first where we linearly transform the input to this block
\begin{align}
    \bm e_t = \bm W_e \bm o_t + \bm b_e ,
\end{align}
and the second performs a single layer MLP transformation as
\begin{align}
    \bm f_t = \texttt{GeLU}(\bm W_f \bm o_t + \bm b_f).
\end{align}
These are then combined through a Hadamard product and linear transformation as
\begin{align}
 \bm e'_t &= \bm e_t \odot \bm f_t, \\    \bm m_t &= \bm W_m \bm e'_t + \bm b_m  .
\end{align}
We then have that the vector $\bm{m}_t$ acts as the output of the residual block given the input $\bm a_t$.

\textbf{Distinction between the Griffin and Hawk models.} Hawk is the more simple of the two architectures proposed in \citep{de2024griffin}. Here, residual blocks using the recurrent block described above are simply stacked on top of each other to form the Hawk architecture.
Griffin, on the other hand, mixes recurrent blocks and local attention. 
In particular, two residual blocks with recurrent blocks are followed by one residual block using local MQA attention \citep{beltagy2020longformer, shazeer2019fast}.

\textbf{Simplifying Assumptions}. We again follow the setup of replacing all \texttt{Sigmoid} and \texttt{tanh} activation functions with \ReLU activations which we denote $\sigma$. Furthermore, we assume for simplicity that $d_\texttt{in} = d_\texttt{hidden}$ by choosing $E=1$. Moreover, the Hawk and Griffin architecture contains residual connections and normalising layers which we omit.\footnote{We will force a lot of our recurrent blocks to implement the identity function. So instead of this, we could implement the $0$ function in the recurrent block and use a residual connection between the residual block input and the output of the recurrent block to achieve the same identity function. However, for clarity we ignore residual connections in our derivations. } We again assume compactness of the input domain $\mathcal{X}$ and denote a vector of finite values $\bm k_{lb}$, such that $\bm k_{lb,i} \leq ( \bm x_t)_i$ for $i \in [d_\texttt{in}]$ and all $\bm x_t \in \mathcal{X}$, just as before. Finally, we assume that $d_{\text{conv}} = T$ where $T$ is the maximum sequence length.

\subsection{Representing the state update using a recurrent block}
\label{sec:Hawk recurrent}
Starting with the input to the Hawk model, which we denote $\bm a_t$, we define this to be a function of the input to the Gated RNN $\bm x_t$ as $\bm a_t = \begin{bmatrix} \bm 0 \\ \bm x_t \\  \end{bmatrix}$.
First, we set $\bm W_a = \bm I$ so that $\bm a_t' = \bm a_t$. Next we choose matrices 
$\tilde{\bm A} =  \begin{bmatrix}
    \bm 0 & \bm A \\ 
    \bm 0 &  \bm 0 \\
\end{bmatrix} $ and $ \tilde{\bm B} =  \begin{bmatrix}
    \bm 0 &  \bm B \\ 
    \bm 0 &  \bm 0 \\
\end{bmatrix}$
which we then use, with a convolutional window size of $d_{\text{conv}}=T$ to form the convolutional kernel $\bm W_M = \begin{bmatrix}
    \tilde{\bm  B},  \tilde{\bm  A} \tilde{ \bm  B}, \tilde{\bm  A}^2 \tilde{\bm  B},  \cdots,  \tilde{\bm  A}^t \bm \tilde{\bm  B}, \cdots 
\end{bmatrix}$. Setting the convolutional bias as $
    \bm b_{\text{conv}} = \begin{bmatrix} \bm 0 \\ \bm 1 
\end{bmatrix}$ gives 
\begin{align} \bm z_t &= \sum_{i=0}^{t-1} \bm W_M[i] \bm M_t[t-i]  ~~+\bm b_\text{conv} ,  \\ &=\tilde{\bm B} \bm a_t + \tilde{\bm A} \tilde{\bm B} \bm a_{t-1} + \cdots + \tilde{\bm A}^{t-1} \tilde{\bm B}\bm a_1 + \begin{bmatrix} \bm 0 \\ \bm 1 
\end{bmatrix} \\ &= \begin{bmatrix}
    \bm s_t \\ \bm 1
\end{bmatrix}.\end{align} 

Now, we pass $\bm z_t$ through the RG-LRU. We set $\Lambda = 0$ so that $\bm a_t = 0$. 
We also define $\bm W_i = \bm 0$ and $\bm b_i = \bm 1$ so that $\bm i_t = \bm 1$. 
This gives us $\bm h_t  = \bm z_t$, so that we pass the output of the one-dimensional convolution through he RG-LRU. 

Next, let's focus on the second branch. 
Making use of the lower bound $\bm k_{lb}$ on the domain $\mathcal{X}$, we set $\bm W_g = \bm I$ and $\bm b_g = \begin{bmatrix}
    \bm 1 \\ -\bm k_{lb}
\end{bmatrix}$
 so that 
 \begin{equation}
     \bm g_t = \sigma \left( \bm I  \begin{bmatrix}
         \bm 0 \\ \bm x_t
     \end{bmatrix} + \begin{bmatrix}
    \bm 1 \\ -\bm k_{lb}
\end{bmatrix} \right)  = \begin{bmatrix}
    \sigma(\bm 1) \\
    \sigma ( \bm x_t - \bm k_{lb})
\end{bmatrix} = \begin{bmatrix}
    \bm 1 \\
    \bm x_t - \bm k_{lb}
\end{bmatrix},
 \end{equation}
 where we used that $(\bm x_t - \bm k_{lb})_i \geq \bm 0$ for every $i$.
 Combining the two branches gives
 \begin{equation}
     \bm h_t' = \begin{bmatrix}
    \bm 1 \\
    \bm x_t - \bm k_{lb}
\end{bmatrix} \odot  \begin{bmatrix}
    \bm s_t \\ \bm 1
\end{bmatrix} = \begin{bmatrix}
    \bm s_t \\ \bm x_t - \bm k_{lb}
\end{bmatrix}.
 \end{equation}
We finally get the output of the recurrent block by defining $\bm W_o = \bm I$ and $\bm b_0 = \begin{bmatrix}
    \bm 0 \\ \bm k_{lb}
\end{bmatrix}$ so that 
\begin{equation} 
\bm o_t = \begin{bmatrix}
    \bm s_t \\ \bm x_t
\end{bmatrix}.
\end{equation}

\subsection{Representing the identity function using a recurrent block}
\label{sec:Hawk recurrent identiy}
We now show that we can pass an input unchanged through a recurrent block. Assume that the input to the recurrent block is $\bm a_t = \begin{bmatrix} \bm a_{1,t} \\ \bm a_{2,t}  \end{bmatrix} $ with $\bm W_a = \bm I$ so that $\bm a_t' = \bm a_t$. Then we define matrices 
$\tilde{\bm A} =  \bm 0 $ and $ \tilde{\bm B} = \bm I $
which we then use to form the convolutional kernel $\bm W_M = \begin{bmatrix}
    \tilde{\bm  B},  \tilde{\bm  A} \tilde{ \bm  B}, \tilde{\bm  A}^2 \tilde{\bm  B},  \cdots,  \tilde{\bm  A}^t \bm \tilde{\bm  B}, \cdots 
\end{bmatrix}$. 
Finally, setting the convolutional bias as $\bm b_{\text{conv}} = \bm 0$ results in $\bm z_t = \bm a_t$. 
From here, we can again set $\Lambda = 0$, $\bm W_i = \bm 0$ and $\bm b_i = \bm 1$ so that $\bm h_t = \bm z_t$. Looking at the second branch and setting $\bm W_g = 0$ and $\bm b_g = \bm 1$ so that $\bm h_t' = \bm h_t$. Finally, we can simply output the input to the recurrent block by setting $\bm W_o = \bm I$ and $\bm b_o = \bm 0$ so that $\bm o_t = \bm h_t$ which means that $\bm o_t = \bm a_t$.     
 
\subsection{Representing each MLP layer as a gated MLP block}
\label{sec:Hawk MLP layer}
We can represent the MLP layers of the networks $\phi(\bm s_t)$ and $\gamma (\bm x_t)$ as described in \cref{eq:gated_linear_rnn} using Gated MLP blocks. We again denote the $i$-th layer of $\phi$ and $\gamma$ as $\phi_i$ and $\gamma_i$. Assume that the input to the gated MLP block is $\bm a_t = \begin{bmatrix} \bm a_{1,t} \\ \bm a_{2,t}  \end{bmatrix} $. 
Then, on the first purely linear branch, let us define $\bm W_e = \bm I$ and $\bm b_e = \bm 1$ so that $\bm e_t = \bm 1$.
On the second non-linear branch, we can define $\bm W_f = \begin{bmatrix}
     \bm W_{\phi_i} &  \bm 0 \\
     \bm 0 & \bm W_{\gamma_i} 
\end{bmatrix}$ and $\bm b_f = \begin{bmatrix}
    \bm b_{\phi_i} \\ \bm b_{\gamma_i}
\end{bmatrix}$. 
This results in 
\begin{equation}
    \bm f_t = 
    \sigma \left( \begin{bmatrix}
     \bm W_{\phi_i} &  \bm 0 \\
     \bm 0 & \bm W_{\gamma_i} 
\end{bmatrix} \begin{bmatrix} \bm a_{1,t} \\ \bm a_{2,t}  \end{bmatrix} + \begin{bmatrix}
    \bm b_{\phi_i} \\ \bm b_{\gamma_i}
\end{bmatrix}  \right) = \begin{bmatrix}
    \phi_i(\bm a_{1, t}) \\
    \gamma_i(\bm a_{2, t})
\end{bmatrix}.
\end{equation}

Due to our setting of $\bm e_t$, we get $\bm e_t ' = \bm f_t$. 
Further, defining $\bm W_m = \bm I$ and $\bm b_m = \bm 0$ makes the output of the MLP block be
\begin{equation}
    \bm m_t =\begin{bmatrix}
    \phi_i(\bm a_{1, t}) \\
    \gamma_i(\bm a_{2, t})
\end{bmatrix}.
\end{equation}

\textbf{Emulating the layers of only one the two networks.} 
Suppose without loss of generality (WLOG) that $\phi $ has $m$ layers and $\gamma$ has $n$ layers where $m < n$. 
Suppose also that our input to the MLP block is $\bm a_t = \begin{bmatrix} \phi (\bm x_t) \\ \bm a_{2,t}  \end{bmatrix} $. 
Again, on the first purely linear branch, let us define $\bm W_e = \bm I$ and $\bm b_e = \bm 1$ so that $\bm e_t = \bm 1$.
Now we modify the weights on the second non-linear branch by defining $\bm W_f = \begin{bmatrix}
     \bm I &  \bm 0 \\
     \bm 0 & \bm W_{\gamma_i} 
\end{bmatrix}$ and $\bm b_f = \begin{bmatrix}
    \bm 0 \\ \bm b_{\gamma_i}
\end{bmatrix}$. This gives us 
\begin{equation}
    \bm f_t = 
    \sigma \left( \begin{bmatrix}
     \bm I &  \bm 0 \\
     \bm 0 & \bm W_{\gamma_i} 
\end{bmatrix} \begin{bmatrix} \phi (\bm x_t) \\ \bm a_{2,t}  \end{bmatrix} + \begin{bmatrix}
    \bm 0 \\ \bm b_{\gamma_i}
\end{bmatrix}  \right) = \begin{bmatrix}
    \sigma(\phi( \bm x_t)) \\
    \gamma_i(\bm a_{2, t})
\end{bmatrix} = \begin{bmatrix}
    \phi( \bm x_t) \\
    \gamma_i(\bm a_{2, t})
\end{bmatrix},
\end{equation}
where we have used that since $\phi(\bm x_t)$ is a ReLU network whose final activation is a ReLU, we have that $\phi(\bm x_t) = \sigma(\phi(\bm x_t))$. Hence, if our networks have different depths and we have fully emulated one of the networks, we can continue to emulate the remaining layers of the other network while keeping the fully emulated network fixed and unchanged.

\subsection{Representing the identify function using a gated MLP block}
\label{sec:Hawk MLP identity}
In this section we show that we can represent an identity function using a gated MLP block. This can be simply done by setting $\bm W_f = \bm 0, \bm b_f = \bm 1, \bm W_e = \bm I, \bm b_e = \bm 0, \bm W_m = \bm I$ and $\bm b_m = \bm 0$. 
This then gives us that for an input  $\bm a_t = \begin{bmatrix} \bm a_{1,t} \\ \bm a_{2,t}  \end{bmatrix} $ to the gated MLP block, the output of the gated MLP block is $\bm m_t = \bm a_t$.
Thus, we pass the input through the gated MLP unchanged.

\subsection{Representing multiplicative gating with a gated MLP block}
\label{sec:Hawk gating}
The final thing we need to do is to compute an element-wise product of two vectors in order to match the output in \cref{eq:gated_linear_rnn}.
In other words, to match the $\phi(\bm x_t) \odot \gamma(\bm s_t)$ operation. 

Again, assume that the input to the gated MLP block is $\bm a_t = \begin{bmatrix} \bm a_{1,t} \\ \bm a_{2,t}  \end{bmatrix}$. Working with the first linear branch, we define $\bm W_e = \begin{bmatrix}  
    \bm 0 & \bm  0 \\
    \bm I & \bm 0
\end{bmatrix}$ and $\bm b_e = \bm 0$, so that 
\begin{equation}
    \bm e_t = 
    \begin{bmatrix}  
    \bm 0 & \bm  0 \\
    \bm I & \bm 0
\end{bmatrix} \begin{bmatrix} \bm a_{1,t} \\ \bm a_{2,t}  \end{bmatrix} + \bm 0  = \begin{bmatrix}
    \bm 0 \\
    \bm a_{1, t}
\end{bmatrix}.
\end{equation}
Next, we define $\bm W_f = \bm I$ and $\bm b_e = \bm 0$ so that 
\begin{equation}
    \bm f_t = \begin{bmatrix}
        \sigma(\bm a_{1,t} ) \\
        \sigma(\bm a_{2,t}) 
    \end{bmatrix}.
\end{equation}
Setting $\bm W_m = \bm I$ and $\bm b_m = \bm 0$ gives the output of the gated MLP as 
\begin{equation}
    \bm m_t = \begin{bmatrix}
        \bm 0 \\
        \bm a_{1, t } \odot \sigma(\bm a_{2,t}) 
    \end{bmatrix}.
\end{equation}

\subsection{Composing the operations to model a single gated linear-RNN layer}
Now that we have all the individual layers, we can combine them so that we can use a Hawk model to emulate a single Gated RNN layer. 

First we start by taking the input of the form $\bm a_t = \begin{bmatrix} \bm 0 \\ \bm x_t \\  \end{bmatrix}$. 
We use a residual block that consists of a recurrent block computing the state update as descried in \cref{sec:Hawk recurrent} and then a gated MLP block that computes the identity function as demonstrated in \cref{sec:Hawk MLP identity}. 
This gives an output from this first recurrent block as $\bm o_t = \begin{bmatrix}
    \bm s_t \\ \bm x_t
\end{bmatrix}$.

Next, we emulate the MLP layers of the networks $\phi$ and $\gamma$ in parallel. Suppose WLOG that $\phi$ and $\gamma$ have $m$ and $n$ MLP layers respectively, where $m \leq n$. We stack $m$ residual blocks using recurrent blocks that implement the identity function as described in \cref{sec:Hawk recurrent identiy} followed by MLP blocks that apply the MLP layers of $\phi$ and $\gamma$ as described in \cref{sec:Hawk MLP layer}. Stacking $m$ such residual blocks results in the output $\bm m_t = \begin{bmatrix} \gamma_m(\bm s_t) \\ \phi(\bm x_t)  \end{bmatrix}$, where we can fully emulate the shallower network $\phi(\bm x_t)$. 

Now, for the remaining $k-m$ layers for the network $\gamma(\bm x_t)$, we stack residual blocks with recurrent blocks implementing the identity function as described in \cref{sec:Hawk recurrent identiy} and MLP blocks that leave $\phi(\bm x_t)$ unchanged whilst applying the additional layers needed to emulate $\gamma(\bm s_t)$ as described at the end of \cref{sec:Hawk MLP layer}. After stacking $k-m$ additional residual layers in this fashion, the output of the final residual block will now be  $\bm m_t = \begin{bmatrix} \gamma(\bm s_t) \\ \phi(\bm x_t)  \end{bmatrix}$, 
 which fully reconstructs the MLP networks $\gamma$ and $\phi$.

Finally, we utilise a residual block with a recurrent block that implements the identity function as described in \cref{sec:Hawk recurrent identiy} followed by a gated MLP block that applies multiplicative gating as described in \cref{sec:Hawk gating}. 
This then gives as an output of this final residual block 
 $\bm m_t = \begin{bmatrix} \bm 0 \\ \gamma(\bm s_t) \odot \sigma(\phi(\bm x_t))  \end{bmatrix}$. Since $\phi(\bm x_t)$ is a MLP network with the final activation function being a ReLU activation, we have that $\sigma(\phi(\bm x_t)) = \phi(\bm x_t)$, giving the required final output from the stacked block of residual blocks as
 \begin{equation}
 \bm m_t = \begin{bmatrix} \bm 0 \\ \gamma(\bm s_t) \odot \phi(\bm x_t)  \end{bmatrix}.
 \end{equation}

Hence, we have shown that a single layer of a gated RNN as described by \cref{eq:gated_rnn} can be represented using $k+2$ Hawk residual blocks where $k $ is the maximum depth of $\phi$ and $\gamma$. Once again, the two universal approximation programs in \Cref{lst:continous_approximation,lst:tok2tok_approximation} can also be applied to Hawk models as they can represent Gated Linear RNNs. 
Therefore, Hawk models are also universal approximators in the sense described in \cref{sec:ua_linear_RNNs}.

\textbf{Gated Linear-RNNs are Griffin models too.} The above argument extends to the Griffin architecture which uses stacks of two residual blocks with recurrent blocks followed by a residual block with attention. The only thing that changes is that for every third residual block, which in our argument will be used to compute the MLP layers of $\phi$ and $\gamma$ in parallel, the recurrent block is now replaced with a local MQA block. 

We can set the key query and values matrices to implement the identity function which is to act input to the block. 
Hence, as a corollary of the above argument, we can also show that the universal approximation programs in \Cref{lst:continous_approximation,lst:tok2tok_approximation} can also be implemented as  Griffin models. 
Therefore, Griffin models can also be universal approximators in the sense described in \cref{sec:ua_linear_RNNs}.

\section{Definitions for some helper functions in LSRL}
\label{sec:sugar_in_lsrl}

\subsection{\texttt{f\_not}}
This is a convenience function that creates a NOT function block.
It assumes that $x$ is 0 or 1.
Works with scalar and vector-valued inputs. 
With vector-valued inputs, it acts element-wise.
\begin{lstlisting}
not_x = 1 - x
\end{lstlisting}

\subsection{\texttt{f\_and}}
This is a convenience function that creates an AND function block.
It assumes that $x$ and $y$ are 0 or 1.
Works with scalar and vector-valued inputs. 
With vector-valued inputs, it acts element-wise.
\texttt{mu} is the approximation parameter $\mu$ for \texttt{f\_step} as described in \Cref{sec:lsrl}.
\begin{lstlisting}
and_x_y = ReLU(f_step(x, mu) + f_step(y, mu) - 1)
\end{lstlisting}

\subsection{\texttt{f\_or}}
This is a convenience function that creates an OR function block.
It assumes that $x$ and $y$ are 0 or 1.
Works with scalar and vector-valued inputs. 
With vector-valued inputs, it acts element-wise.
\texttt{mu} is the approximation parameter $\mu$ for \texttt{f\_step} as described in \Cref{sec:lsrl}.
\begin{lstlisting}
or_x_y = f_step(x + y, mu=mu)
\end{lstlisting}

\subsection{\texttt{f\_smaller}}
This is a convenience function that a less than comparison block.
Works with scalar and vector-valued inputs. 
With vector-valued inputs, it acts element-wise.
\texttt{mu} is the approximation parameter $\mu$ for \texttt{f\_step} as described in \Cref{sec:lsrl}.
\begin{lstlisting}
smaller_x_y = f_step(y - x, mu=mu)
\end{lstlisting}

\subsection{\texttt{f\_larger}}
This is a convenience function that a more than comparison block.
Works with scalar and vector-valued inputs. 
With vector-valued inputs, it acts element-wise.
\texttt{mu} is the approximation parameter $\mu$ for \texttt{f\_step} as described in \Cref{sec:lsrl}.
\begin{lstlisting}
larger_x_y = f_step(x - y, mu=mu)
\end{lstlisting}

\subsection{\texttt{f\_relu\_identity}}
\label{sec:relu_identity}
Identity operation using \ReLU{}s.
This is useful for debranching when some of the branches have \ReLU{}s but the other don't.
We can add this as a bypass for the ones that do not and can then merge the \ReLU{}s together (see \Cref{sec:debranching_rules} for details).
\begin{lstlisting}
positive_part = ReLU(x)
negative_part = ReLU(
    Linear(
        input=x,
        A=-1 * eye(x.dim),
        b=zeros(x.dim, 1),
    )
)
both = Concat([positive_part, negative_part])
relu_identity = Linear(
    input=both,
    A=hstack(eye(x.dim), -1 * eye(x.dim)),
    b=zeros(x.dim, 1),
)
\end{lstlisting}

\subsection{\texttt{f\_modulo\_counter}}
\label{sec:f_modulo_counter}
Computes the $x \texttt{ mod } \texttt{divisor}$ where $x$ is a counter starting from zero.
The idea is that we rotate a unit vector so that it makes a full revolution every \texttt{divisor} rotations.
\texttt{dummy\_input} can be any variable, we use it only to construct a constant.

\begin{lstlisting}
angle = 2 * pi / divisor
R = [[cos(angle), sin(angle)], [sin(angle), cos(angle)]]

unit_vector = [[1], [0]]

# we first rotate, then output so if we want the first output to be 0 we need to have the init_state one step before that
init_state = R.inv() @ unit_vector

# this rotates a 2D vector 1/divisor revolutions at a time
cycler = LinState(
    input=dummy_input,
    A=R,
    B=zeros(2, dummy_input.dim),
    init_state=init_state,
)

# we now need to extract the position of the cycler
extractor_matrix = vstack(*[(R^i * unit_vector).T) for i in range(divisor)])
indicator = Linear(
    input=cycler, 
    A=extractor_matrix, 
    b=zeros(divisor, 1)
)

# the dot product with the row of extractor_matrix corresponding to the current position of the cycler is 1
# the dot product with the second highest is cos(angle)
# thus, we can threshold at 1-cos(angle/2) to get a one hot encoding of the current position of the cycler
one_hot = f_larger(indicator, cos(angle / 2))

# and to get an integer value we need one final linear layer
mod_value = Linear(
    one_hot, 
    A=[[i for i in range(divisor)]], 
    b=zeros(1, 1)
)
\end{lstlisting}

\newpage

\begin{taggedblock}{neurips}
\section*{NeurIPS Paper Checklist}

\begin{enumerate}

\item {\bf Claims}
    \item[] Question: Do the main claims made in the abstract and introduction accurately reflect the paper's contributions and scope?
    \item[] Answer: \answerYes{}
    \item[] Justification: The abstract and the introduction clearly state all the contributions of the paper and clearly differentiate the theoretical results which hold in general and the empirical phenomena that we observe, which may not generalize to all settings.

\item {\bf Limitations}
    \item[] Question: Does the paper discuss the limitations of the work performed by the authors?
    \item[] Answer: \answerYes{}
    \item[] Justification: The paper discusses the limitations of the present work. There is a dedicated \emph{Limitations} section in \Cref{sec:discussion} that addresses the fact that we only provide constructive existence results but not necessary and sufficient conditions for universal in-context approximation to arise. We also highlight that our results might not hold to models with structural constraints on their parameters. Moreover, we have a dedicated section (\Cref{sec:ua_gated_linear_RNNs}) which addresses some of the limitations of constructing universal in-context approximators with fully recurrent architectures in practice. This section proposes solutions and demonstrates that they result in more numerically stable models which are more likely to occur in practice.

\item {\bf Theory Assumptions and Proofs}
    \item[] Question: For each theoretical result, does the paper provide the full set of assumptions and a complete (and correct) proof?
    \item[] Answer: \answerYes{}
    \item[] Justification: The paper has two main theoretical results: the constructions of universal in-context approximators for continuous and for discrete functions. Both results are presented as LSRL programs which compile to the architectures considered in this work. Furthermore, these programs have been implemented in Python, their correctness has been tested and they are available in the supplementary materials. 

    \item {\bf Experimental Result Reproducibility}
    \item[] Question: Does the paper fully disclose all the information needed to reproduce the main experimental results of the paper to the extent that it affects the main claims and/or conclusions of the paper (regardless of whether the code and data are provided or not)?
    \item[] Answer: \answerYes{}
    \item[] Justification: There are two experimental aspects to this work.
        First, there is the implementation of LSRL and the two universal approximation programs in \Cref{lst:continous_approximation,lst:tok2tok_approximation}.
        The most critical aspect of implementing LSRL is the debranching algorithm which is described in detail in \Cref{sec:debranching_rules}. Additionally, the two programs are described in full in their corresponding listings. We also provide Python implementation for the LSRL compiler and the two programs.

        Second, there is the study of how affected by parameter noise are the different implementations of the conditional assignment operator \texttt{f\_ifelse} which was presented in \Cref{sec:ua_nonlinear_RNNs}.
        The details of this experiment are described in \Cref{fig:ifelse_noise} and we also provide the code with which we did the experiment and our plots.

\item {\bf Open access to data and code}
    \item[] Question: Does the paper provide open access to the data and code, with sufficient instructions to faithfully reproduce the main experimental results, as described in supplemental material?
    \item[] Answer: \answerYes{}
    \item[] Justification: We are providing code that includes an implementation of LSRL, the two universal in-context approximation programs in \Cref{lst:continous_approximation,lst:tok2tok_approximation} and everything needed to reproduce the experiments in this work.

\item {\bf Experimental Setting/Details}
    \item[] Question: Does the paper specify all the training and test details (e.g., data splits, hyperparameters, how they were chosen, type of optimizer, etc.) necessary to understand the results?
    \item[] Answer: \answerNA{}
    \item[] Justification: The experiments in our work are based on \emph{constructed} models rather than \emph{trained} models. Therefore, considerations such as dataset, optimizers and hyperparameters do not apply.

\item {\bf Experiment Statistical Significance}
    \item[] Question: Does the paper report error bars suitably and correctly defined or other appropriate information about the statistical significance of the experiments?
    \item[] Answer:  \answerNo{}
    \item[] Justification: For this work, uncertainty quantification could only make sense in the context of \Cref{fig:ifelse_noise}. 
        However, the behaviour we observe, especially for the continuous case, is bimodal.
        As bimodal distributions cannot be properly captured with error bars we decided against using them.
        Furthermore, we are studying whether a phenomenon occurs, rather than quantifying it.
        Therefore, we decided to instead use a strip plot instead as it explicitly shows all our results unabridged, clearly indicates the bimodal nature of the results, and distinctly showcases the noise robustness trends of the different approaches we consider.

\item {\bf Experiments Compute Resources}
    \item[] Question: For each experiment, does the paper provide sufficient information on the computer resources (type of compute workers, memory, time of execution) needed to reproduce the experiments?
    \item[] Answer: \answerNo{}
    \item[] Justification: Our experiments were ran on a single machine and using only CPU compute. Therefore, the compute required is negligible for the contemporary machine learning standards.
    
\item {\bf Code Of Ethics}
    \item[] Question: Does the research conducted in the paper conform, in every respect, with the NeurIPS Code of Ethics \url{https://neurips.cc/public/EthicsGuidelines}?
    \item[] Answer: \answerYes{}
    \item[] Justification: This is a theoretical work with no human participants, datasets, or potential societal impact or harmful consequences. Therefore, the present work has no  moral or ethical relevance or implications.

\item {\bf Broader Impacts}
    \item[] Question: Does the paper discuss both potential positive societal impacts and negative societal impacts of the work performed?
    \item[] Answer: \answerNA{}
    \item[] Justification: As mentioned above, this is a theoretical work which establishes theoretical properties of mathematical objects that are already used in practice. However, we do discuss the implications of our findings, namely that if models are universal in-context approximators, then it might be difficult to ensure that they are aligned and cannot be misused. Nevertheless, we only show that this is a property already present in existing models, and hence our work does not introduce new attack or misuse vectors. On the contrary, we hope that us highlighting this issues will help the community to develop safer and more secure generative AI systems.
    
\item {\bf Safeguards}
    \item[] Question: Does the paper describe safeguards that have been put in place for responsible release of data or models that have a high risk for misuse (e.g., pretrained language models, image generators, or scraped datasets)?
    \item[] Answer: \answerNA{}
    \item[] Justification: We release no data or models.

\item {\bf Licenses for existing assets}
    \item[] Question: Are the creators or original owners of assets (e.g., code, data, models), used in the paper, properly credited and are the license and terms of use explicitly mentioned and properly respected?
    \item[] Answer: \answerNA{}
    \item[] Justification: The paper does not use existing assets beyond common Python libraries.

\item {\bf New Assets}
    \item[] Question: Are new assets introduced in the paper well documented and is the documentation provided alongside the assets?
    \item[] Answer: \answerYes{}
    \item[] Justification: The only new asset arising from this work is the LSRL code base which we have ensured to be well-documented, accompanied by unit and integration tests, and with illustrative Jupyter notebooks. 

\item {\bf Crowdsourcing and Research with Human Subjects}
    \item[] Question: For crowdsourcing experiments and research with human subjects, does the paper include the full text of instructions given to participants and screenshots, if applicable, as well as details about compensation (if any)? 
    \item[] Answer: \answerNA{}
    \item[] Justification: There were no human participants involved in any part of this work.

\item {\bf Institutional Review Board (IRB) Approvals or Equivalent for Research with Human Subjects}
    \item[] Question: Does the paper describe potential risks incurred by study participants, whether such risks were disclosed to the subjects, and whether Institutional Review Board (IRB) approvals (or an equivalent approval/review based on the requirements of your country or institution) were obtained?
    \item[] Answer: \answerNA
    \item[] Justification: This is a purely theoretical work and as such no IRB approval or equivalent was necessary or appropriate.

\end{enumerate}
\end{taggedblock}

\end{document}